\documentclass[a4]{article}
\usepackage{hyperref}
\usepackage{url}
\usepackage[T1]{fontenc}
\usepackage{ae,aecompl}
\usepackage[utf8x]{inputenc}
\usepackage{amsmath}
\usepackage{empheq}
\usepackage{amsfonts}
\usepackage{amssymb}
\usepackage{amsthm}
\usepackage{graphicx, textcomp, caption}
\usepackage{pst-math}
\usepackage{pst-node}
\usepackage[affil-it]{authblk}

\usepackage{placeins}

\let\Oldsection\section
\renewcommand{\section}{\FloatBarrier\Oldsection}

\let\Oldsubsection\subsection
\renewcommand{\subsection}{\FloatBarrier\Oldsubsection}

\let\Oldsubsubsection\subsubsection
\renewcommand{\subsubsection}{\FloatBarrier\Oldsubsubsection}

\title{PFAx: Predictable Feature Analysis to Perform Control}

\author{Stefan~Richthofer\footnote{Electronic address: \texttt{stefan.richthofer@ini.rub.de}; 
Corresponding author}, Laurenz~Wiskott\footnote{Electronic address: 
\texttt{laurenz.wiskott@ini.rub.de}}}
\affil{Institut f\"ur Neuroinformatik,\\ Ruhr-Universit\"at Bochum, Germany}

\providecommand{\abs}[1]{\lvert#1\rvert}

\providecommand{\norm}[1]{\lVert#1\rVert}

\providecommand{\av}[1]{\left\langle#1\right\rangle}

\providecommand{\bigav}[1]{\big\langle#1\big\rangle}

\providecommand{\coloneq}[0]{\mathrel{\mathop:}=}
\providecommand{\id}[0]{\mathbf{I}}
\providecommand{\trph}[0]{\Omega_t}
\providecommand{\eqcolon}[0]{= \mathrel{\mathop:}}

\DeclareMathOperator*{\eq}{=}
\DeclareMathOperator*{\appr}{\approx}

\DeclareMathOperator{\mvec}{vec}
\DeclareMathOperator{\hist}{hist}

\DeclareMathOperator{\tr}{Tr}

\DeclareMathOperator{\orth}{O}

\DeclareMathOperator*{\opmin}{minimize}

\providecommand{\optmin}[1]{\displaystyle\opmin_{#1} \qquad}
\DeclareMathOperator*{\subjectto}{subject \; to \qquad}

\providecommand{\pfaerrob}[2]{\bigav{\; \norm{#2-#1}^2 \; }}

\newtheoremstyle{defi}
  {}
  {}
  {\slshape}
  {}
  {\bfseries}
  {}
  {\newline}
  {}

\theoremstyle{defi}

\begin{document}

\maketitle

\begin{abstract}
Predictable Feature Analysis (PFA) \cite{DBLP:conf/icmla/RichthoferW15} is an algorithm that performs dimensionality reduction on high dimensional input signal. It extracts those subsignals that are most predictable according to a certain prediction model. We refer to these extracted signals as \textsl{predictable features}.

In this work we extend the notion of PFA to take supplementary information into account for improving its predictions. Such information can be a multidimensional signal like the main input to PFA, but is regarded external. That means it won't participate in the feature extraction -- no features get extracted or composed of it. Features will be exclusively extracted from the main input such that they are most predictable based on themselves and the supplementary information.
We refer to this enhanced PFA as PFAx (PFA extended).

Even more important than improving prediction quality is to observe the effect of supplementary information on feature selection. PFAx transparently provides insight how the supplementary information adds to prediction quality and whether it is valuable at all.
Finally we show how to invert that relation and can generate the supplementary information such that it would yield a certain desired outcome of the main signal.

We apply this to a setting inspired by reinforcement learning and let the algorithm learn how to control an agent in an environment. With this method it is feasible to locally optimize the agent's state, i.e.\ reach a certain goal that is near enough. We are preparing a follow-up paper that extends this method such that also global optimization is feasible.
\end{abstract}

\pagebreak

\section{Introduction}
The reinforcement learning (RL) setting consists of an agent in an environment and usually involves three signals -- perception, action and reward. The agent has access to the perception and reward signal and can generate the action signal as an output, which in turn yields consequences for future perception and reward. The agent's goal is to maximize reward over a certain time period.

\begin{figure}[h!]
	\centering
	\includegraphics[width=0.6\hsize]{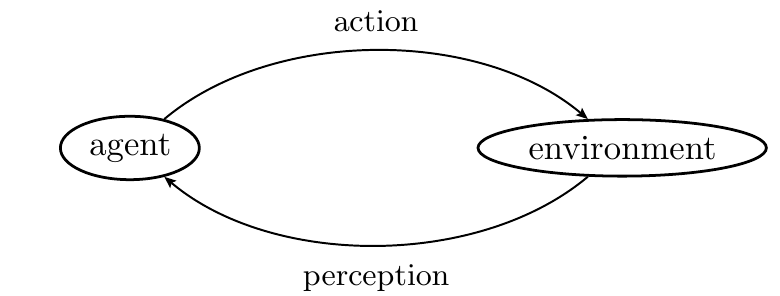}
%
%
	\caption{Perception/control cycle}
	\label{fig:cycle}
\end{figure}

In this work we focus on the action/perception cycle (figure \ref{fig:cycle}) of RL scenarios with continuous state space, perception signal and action signal. We -- however -- do not consider an arbitrary reward signal for now, but just aim for reaching specific states indicated by their corresponding perception.
A core issue of reinforcement learning is to represent the agent's state in a way that is of tractable complexity and at the same time allows to perform the desired task. In \cite{DBLP:conf/icmla/RichthoferW15} we have presented Predictable Feature Analysis (PFA) as an algorithm to focus on features of the environment that are intrinsically predictable. The idea is that predictability is a crucial property for any task -- without it, one cannot estimate consequences of possible actions and would have to act randomly. There are -- however -- different notions of predictability, e.g.\ differing in the choice of model, whether a model is used at all, what to predict and what a prediction can be based on. These factors can crucially make a difference in how useful predictability really is for an RL scenario. The original PFA approach selects features that are most suitable to predict themselves.

We go one step further now and propose an extension to PFA that enables more fine-grained control of what a prediction can be based on.
This allows to incorporate the action or control signal of RL settings and thus to aim for features that are not only predictable, but that can be well manipulated.
To accomplish this, we extend PFA to PFAx by enabling it to take any supplementary information into account for prediction. Note that such supplementary information will not be used for data extraction -- PFAx will compose features exclusively from its main input such that they are well predictable based on themselves and given the supplementary information. While we have a concrete benefit for RL in mind, this enriches the approach and opens up various new applications more generally.

Beneath improving prediction quality, this enhancement enables us to study the effect of supplementary information on feature selection. PFAx provides a coefficient matrix that transparently shows how valuable the supplementary signal is for prediction of the extracted features.
It turns out that the prediction rule is invertible such that we can generate an action signal that is most likely to result in a specific desired manipulation given the agent's current state. This way we can perform goal directed navigation in an RL environment.

\subsection{Related work}
A related SFA-inspired approach for RL is Contingent Feature Analysis (CFA) \cite{DBLP:conf/icann/Sprague14}. The main difference to the PFAx based approach presented here is that CFA assumes a discrete action set and uses a different selection criterion. PFAx assumes a continuous signal for control -- think of pressing buttons (discrete control) compared to turning a steering wheel or moving levers (continuous control). CFA requires a no-operation (NOP) among the possible actions and then selects features to behave as uncorrelated as possible to NOP induced behavior.
Given that continuous control is discretely sampled in digital processing, a comparison of PFAx and CFA would be an interesting exercise in the future, but is not in the scope of this work.

There are a number of approaches with the same goal as described in \cite{DBLP:conf/icmla/RichthoferW15}.

Forecastable Component Analysis (ForeCA) \cite{goerg13}, is a method based on the same paradigm as PFA, but using a model-independent approach. There are also differences in scalability and we investigated a side-by-side comparison of PFA, ForeCA and SFA in \cite{DBLP:conf/icmla/RichthoferW15} to some extend.

Graph-based Predictable Feature Analysis (GPFA) \cite{Weghenkel:2017:GPF:3140707.3140724} analyses the distribution of data samples given previous samples. Low variance in that distribution is used to detect predictability. The work draws links to graph embedding and an information-theoretic measure of predictive information. It compares GPFA with SFA, PFA and ForeCA under various criteria, including prediction quality and runtime performance.

Predictive Projections \cite{DBLP:conf/ijcai/Sprague09} is an approach inspired by metric learning, more specifically by Neighborhood Components Analysis (NCA) \cite{DBLP:conf/nips/GoldbergerRHS04}. It uses conjugate gradient descent to find good projections for accurate prediction of future states in Markov Decision Processes. While PFA is motivated by a comparable goal, it uses a different problem formulation and  optimization technique.

In \cite{boxTiao1977} a comparable approach to PFA is formulated and PFA can be seen as a generalization of the method presented there. We will refer to that method again in a later section.

\section{Extracting predictable features} \label{sec:extraction}
We start with a comprehension of the PFA algorithm \cite{DBLP:conf/icmla/RichthoferW15} and then extend the method and notation to incorporate supplementary information. Given an input-signal $\mathbf{x}(t)$ with $n$ components, PFA's objective is to find $r$ most predictable output components, referred to as “predictable features”.

The extraction itself is linear but can be enhanced by a non-linear expansion $\mathbf{h}$. Expanding the input by monomials up to a desired degree, the extraction function essentially becomes a polynomial of that degree\footnote[1]{For high-degree expansion, Legendre or Bernstein polynomials are preferable bases compared to monomials because of better numerical stability.}. Note that by the Stone-Weierstrass theorem this technique can approximate any continuous function. This further covers regulated functions (i.e.\ piecewise continuous) as these can be approximated by continuous functions. However, raising the expansion to high degree might require a lot of training data and cause high computational costs. Applying PFA in a hierarchical manner like is done with SFA in \cite{FranziusSprekelerEtAl-2007e, FranziusSprekelerEtAl-2007d, Schoenfeld2015} can help to keep computation tractable.

As a prerequisite for PFA we shortly recall the SFA algorithm.

\subsection{Recall SFA} \label{sec:sfa}
Instead of predictability, SFA optimizes for slow variation. Extraction is performed by linear transformation and projection. The extraction matrix is optimized over a finite training-phase $\trph$ consisting of equidistant time points. Like mentioned above, a non-linear expansion $\mathbf{h}$ can be applied to the signal.
To avoid trivial constant solutions, output is constrained to have unit variance and zero mean. Further more, the output signals must be pairwise uncorrelated to avoid redundant output components.
Mean is defined as $\av{s(t)}_t~\coloneq~\frac{1}{\abs{\trph}} \sum_{t\in \trph} s(t)$ (average over training phase). The initial step of SFA is to sphere the expanded signal over the training-phase, i.e.\ shift its mean to zero and normalize the covariance-matrix to identity:
\begin{align} \label{sphering}
\tilde{\mathbf{z}}(t) \quad &\coloneq \quad \mathbf{h}(\mathbf{x}(t)) - \av{\mathbf{h}(\mathbf{x}(t))}_t &&\text{(make mean-free)}\\
\mathbf{z}(t) \quad &\coloneq \quad \mathbf{S} \tilde{\mathbf{z}}(t) \qquad \text{with} \quad \mathbf{S} \coloneq \av{\tilde{\mathbf{z}} \tilde{\mathbf{z}}^T}^{-\frac{1}{2}} &&\text{(normalize covariance)}
\end{align}
SFA then becomes the following optimization problem:
\begin{align} \label{sfa}
\text{For} \; i \in \{1, \ldots, r\} \notag \\
	\begin{split} 
		\optmin{\mathbf{a}_i \in \mathbb{R}^{n}} & \mathbf{a}_i^T \av{\dot{\mathbf{z}}\dot{\mathbf{z}}^T} \mathbf{a}_i\\
		\subjectto	& \mathbf{a}_i^T \av{\mathbf{z}} \hphantom{\mathbf{a}_i \mathbf{a}_j \mathbf{z}^T} \, = \quad 0 \quad \hphantom{\forall \; j < i} \quad \text{(zero mean)}\\
				& \mathbf{a}_i^T \av{\mathbf{z} \mathbf{z}^T} \mathbf{a}_i \hphantom{\mathbf{a}_j} = \quad 1 \quad \hphantom{\forall \; j < i} \quad \text{(unit variance)}\\
				& \mathbf{a}_i^T \av{\mathbf{z} \mathbf{z}^T} \mathbf{a}_j \hphantom{\mathbf{a}_i} = \quad  0 \quad \forall \; j < i \quad \text{(pairwise uncorrelated)}
	\end{split}
\end{align}
Because of sphering it holds that $\av{\mathbf{z}} = 0$ and $\av{\mathbf{z} \mathbf{z}^T} = \id$, transforming the constraints to
\begin{equation} \label{SFA-constrains-no-matrix}
\mathbf{a}_i^T \mathbf{a}_j = \delta_{ij}
\end{equation}
For later equations a matrix notation of this constraint will be handy. With $\mathbf{A}_r~\coloneq~\left(\mathbf{a}_1,~\ldots,~\mathbf{a}_r~\right)~\in~\mathbb{R}^{n~\times~r}$ denoting the extraction matrix, \eqref{SFA-constrains-no-matrix} is equal to
\begin{equation} \label{SFA-constrains-matrix}
\exists \mathbf{A} \in \orth(n) \colon \qquad \mathbf{A}_r \quad = \quad \mathbf{A}\mathbf{I}_r
\end{equation}
$\orth(n) \subset \mathbb{R}^{n \times n}$ denotes the space of orthogonal transformations, i.e.\ $\mathbf{A}\mathbf{A}^T = \mathbf{I}$ and $\mathbf{I}_r \in \mathbb{R}^{n \times r}$ denotes the reduced identity matrix consisting of the first $r$ Euclidean unit vectors as columns.
Choosing $\mathbf{a}_i$ as eigenvectors of $\av{\dot{\mathbf{z}}\dot{\mathbf{z}}^T}$, corresponding to the eigenvalues in
ascending order, yields $\mathbf{A}_r$ solving \eqref{sfa} globally. \cite{WiskottBerkesEtAl-2011} describes this procedure in detail. In the following $\mathbf{m} \coloneq \mathbf{A}_r^T \mathbf{z}$ denotes the extracted signal.

\subsection{Modeling the PFA-problem} \label{sec:pfa}
To measure predictability, we focus on \textbf{linear, auto-regressive prediction} as our model -- it is successfully used to model various time-related problems.
In this model, a signal is predictable if it can be approximated by a linear combination of $p$ recent values.
Formally, this yields the problem of finding vectors $\mathbf{a}$ and $\mathbf{b}$ such that
\begin{align}
 \mathbf{a}^T \mathbf{z}(t) \quad \appr^! \;& \quad b_{1} \mathbf{a}^T \mathbf{z}(t-1) + \ldots + b_{p} \mathbf{a}^T \mathbf{z}(t-p) \label{pfa-criterionARScalar}\\
= \;& \quad \mathbf{a}^T \hist_{\mathbf{z}, p}(t) \; \mathbf{b}
\end{align}
with $\hist$ defined as the signal's history of $p$ time-steps:
\begin{equation}
\label{history}
	\hist_{\mathbf{z}, p, \Delta}(t) \quad \coloneq \quad \sum_{i=1}^p \quad \mathbf{z}(t-i\Delta) \mathbf{e}_i^T \quad \text{with} \quad \mathbf{e}_i \in \mathbb{R}^p, \quad \left(\mathbf{e}_1, \ldots, \mathbf{e}_p \right) = \mathbf{I}_{p, p}.
\end{equation}
$\mathbf{I}_{p, p}$ denotes the $p$-dimensional identity and $\mathbf{e}_i$ denotes the $i$-th $p$-dimensional Euclidean unit vector. $\Delta$ defaults to $1$: $\hist_{\mathbf{z}, p} \coloneq \hist_{\mathbf{z}, p, 1}$.

\begin{figure}[ht]
   \centering
   \captionsetup{width=.7\linewidth}
   \includegraphics[width=0.75\hsize]{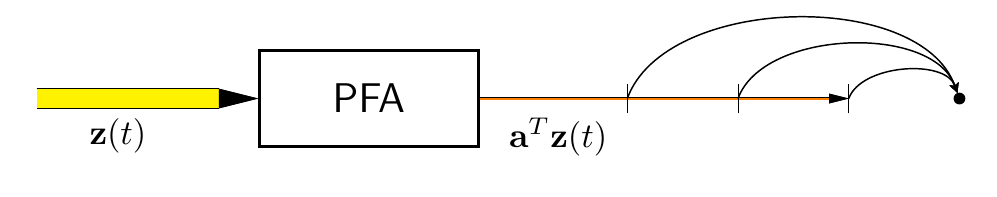}
%
%
%
%

   \caption{Illustration of PFA. Components are selected to be well predictable based on themselves.}
   \label{fig:PFAIllustration}
 \end{figure}

Like in SFA, we optimize parameters over $\trph$ and also adopt the constraints to avoid trivial or redundant solutions. The first steps of PFA are indeed equal to those in SFA, i.e.\ non-linear expansion and sphering. Where possible, we adopt notation from \ref{sec:sfa}.
Commonly, \eqref{pfa-criterionARScalar} is extended to multiple dimensions as follows:
\begin{equation} \label{pfa-criterionARDiagMatrix}
 \mathbf{m}(t) \quad \appr^! \quad \mathbf{B}_{1} \mathbf{m}(t-1) + \ldots + \mathbf{B}_{p} \mathbf{m}(t-p) \qquad \text{with}\quad \mathbf{B}_i \in \mathbb{R}^{n \times n} \text{, diagonal}
\end{equation}
In \cite{DBLP:conf/icmla/RichthoferW15} we explain why this restriction to diagonal $\mathbf{B}_i$ is not suitable for PFA. We generalize it to:
\begin{align} \label{pfa-criterionARGeneralMatrix}
 \mathbf{m}(t) \quad \appr^! &\quad \mathbf{B}_{1} \mathbf{m}(t-1) + \ldots + \mathbf{B}_{p} \mathbf{m}(t-p) \qquad \text{with}\quad \mathbf{B}_i \in \mathbb{R}^{n \times n} \\ 
                     = &\quad \mathbf{B} \mvec(\hist_{\mathbf{z}, p}(t)) \qquad \text{with}\quad \mathbf{B} \; \coloneq \; \left( \mathbf{B}_{1}, \ldots, \mathbf{B}_{p} \right) \; \in \; \mathbb{R}^{n \times np}
\end{align}
In that formulation, each extracted component's prediction can utilize all other extracted components. \eqref{pfa-criterionARDiagMatrix} and \eqref{pfa-criterionARGeneralMatrix} are equal for $n = 1$. 
For \eqref{pfa-criterionARGeneralMatrix} we can initially fit our data in full dimension and search for the best-fitted components afterwards. This would not be possible for \eqref{pfa-criterionARDiagMatrix}, because fitting quality of each component is not invariant under the transformation used for extraction.  Nevertheless, we mention strategies to solve \eqref{pfa-criterionARDiagMatrix} in the appendix, section \ref{sec:extractIsolated}.

Analytically, we obtain the following regression formula for an optimal $\mathbf{B}$, given an orthogonal extraction matrix $\mathbf{A}_r \, \in \, \mathbb{R}^{n \times r}$:
\begin{equation} \label{B-ReducedDimRankTransformSolution}
 \mathbf{B}_{\mathbf{z}}(\mathbf{A}_r) \quad \coloneq \quad \mathbf{A}^T_r \underbrace{\av{\mathbf{z}\mathbf{\zeta}^T}}_{n \times np}\underline{\mathbf{A}_r} \Big(\underline{\mathbf{A}_r}^T\underbrace{\av{\mathbf{\zeta}\mathbf{\zeta}^T}}_{np \times np} \underline{\mathbf{A}_r} \Big)^{-1} \quad \in \; \mathbb{R}^{r \times rp}
\end{equation}
In \eqref{B-ReducedDimRankTransformSolution} we used $\zeta(t) \coloneq \mvec(\hist_{\mathbf{z}, p}(t)) \; \in \; \mathbb{R}^{np}$ and the following shortcut notation defined for any matrix $\mathbf{M} \; \in \; \mathbb{R}^{n \times m}$:
\begin{equation} \label{multiA}
\underline{\mathbf{M}} \quad \coloneq \quad \mathbf{I}_{p, p} \otimes \mathbf{M} \quad = \qquad \underbrace{\!\!\!\!\!\!\left( \begin{smallmatrix} \mathbf{M}&  & \mathbf{0} \\  & \ddots &  \\ \mathbf{0} &  & \mathbf{M} \end{smallmatrix} \right)\!\!\!\!\!\!}_{\text{$p$ times $\mathbf{M}$}} \qquad \in \; \mathbb{R}^{np \times mp}
\end{equation}

Equation \eqref{B-ReducedDimRankTransformSolution} can be derived as follows. For a given $\mathbf{A}_r$ the optimal $\mathbf{B}$ must solve
\begin{equation} \label{B-ReducedDimRankTransformSolution-optProblem}
\optmin{\mathbf{B} \in \mathbb{R}^{r \times rp}} \;
\pfaerrob{\mathbf{B} \underline{\mathbf{A}_r}^T \zeta}{\mathbf{A}_r^T \mathbf{z}} \quad \eqcolon \; f(\mathbf{B})
\end{equation}
We expand the value function $f(\mathbf{B})$
\begin{equation} \label{B-ReducedDimRankTransformSolution-expand}
f(\mathbf{B}) \quad = \quad \mathbf{A}_r^T \av{\mathbf{z} \mathbf{z}^T} \mathbf{A}_r -  \mathbf{A}_r^T \av{\mathbf{z} \mathbf{\zeta}^T} \underline{\mathbf{A}_r} \mathbf{B}^T -  \mathbf{B} \underline{\mathbf{A}_r}^T \av{\mathbf{\zeta} \mathbf{z}^T} \mathbf{A}_r +  \mathbf{B} \underline{\mathbf{A}_r}^T \av{\mathbf{\zeta} \mathbf{\zeta}^T} \underline{\mathbf{A}}_r \mathbf{B}^T
\end{equation}
and set its matrix derivative to zero
\begin{equation} \label{B-ReducedDimRankTransformSolution-derive}
\frac{\partial}{\partial \mathbf{B}} f(\mathbf{B}) \quad = \quad -2 \mathbf{A}_r^T \av{\mathbf{z} \mathbf{\zeta}^T} \underline{\mathbf{A}_r} + 2 \mathbf{B} \underline{\mathbf{A}_r}^T \av{\mathbf{\zeta} \mathbf{\zeta}^T} \underline{\mathbf{A}}_r \quad \eq^! \quad \mathbf{0}
\end{equation}
Solving \eqref{B-ReducedDimRankTransformSolution-derive} for $\mathbf{B}$ yields \eqref{B-ReducedDimRankTransformSolution}.

We define $\mathbf{W}$ to denote the optimal prediction matrix for the sphered signal $\mathbf{z}$ without any extraction or transformation applied yet, i.e.\ set $\mathbf{A}_r = \mathbf{I}$ in \eqref{B-ReducedDimRankTransformSolution}:
\begin{equation} \label{W}
\mathbf{W} \quad \coloneq \quad \mathbf{B}_{\mathbf{z}}(\mathbf{I}) \quad = \quad \av{\mathbf{z}\mathbf{\zeta}^T} \av{\mathbf{\zeta}\mathbf{\zeta}^T}^{-1} \quad \in \; \mathbb{R}^{n \times np}
\end{equation}

It can happen that $\av{\mathbf{\zeta}\mathbf{\zeta}^T}$ is not (cleanly) invertible due to very small or zero-valued eigenvalues. We regard it best practice to project away the eigenspaces corresponding to eigenvalues below a critical threshold. These indicate redundancies in the signal and should not be used for prediction. In an eigenvalue decomposition of $\av{\mathbf{\zeta}\mathbf{\zeta}^T}$ replace eigenvalues below the threshold by $0$ and invert the others. Use the resulting matrix as a proxy for $\av{\mathbf{\zeta}\mathbf{\zeta}^T}^{-1}$.
For $r=n$ and $\mathbf{A} \in \orth(n)$, we have $\mathbf{B}_{\mathbf{z}}(\mathbf{A}) = \mathbf{A}^T\mathbf{W}\underline{\mathbf{A}} \,\in \, \mathbb{R}^{n \times np}$.

If $\mathbf{z}$ is sphered, PFA yields the following problem:
\begin{equation} \label{pfaAutoRegressiveWhitened}
\optmin{\mathbf{A} \in \orth(n)} \;
\pfaerrob{\mathbf{B}_{\mathbf{z}}(\mathbf{A}_r) \underline{\mathbf{A}_r}^T \zeta}{ \mathbf{A}_r^T \mathbf{z}}
\end{equation}
Projection composed with inversion renders \eqref{pfaAutoRegressiveWhitened} intractable by every method known to us (not counting local or probabilistic methods). We propose the following tractable relaxation:
\begin{equation} \label{pfaNonAutoRegressiveWhitened}
\optmin{\mathbf{A} \in \orth(n)} \; \pfaerrob{\mathbf{I}_r^T\mathbf{B}_{\mathbf{z}}(\mathbf{A}) \underline{\mathbf{A}}^T \zeta}{ \mathbf{A}_r^T \mathbf{z}} \quad = \quad \bigav{\; \norm{\mathbf{A}_r^T(\mathbf{z}-\mathbf{W}\mathbf{\zeta})}^2 \; }
\end{equation}

Informally, \eqref{pfaNonAutoRegressiveWhitened} aims for components predictable based on the entire input, while \eqref{pfaAutoRegressiveWhitened} aims for components predictable based on themselves.
Let  $\mathbf{A}_r^*$ denote a global optimum of \eqref{pfaAutoRegressiveWhitened} and $\mathbf{A}_r^{(0)}$ a global optimum of \eqref{pfaNonAutoRegressiveWhitened}.

\begin{figure}[ht]
   \centering
   \captionsetup{width=.7\linewidth}
   \includegraphics[width=0.75\hsize]{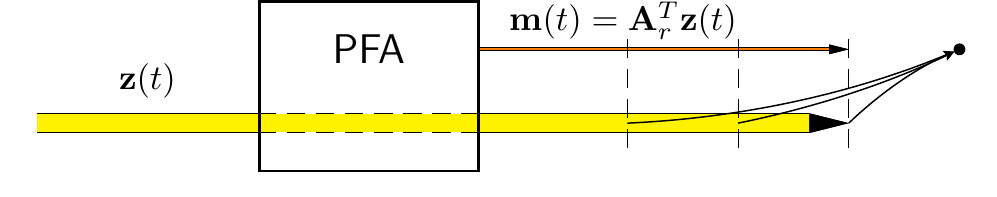}
%
%
%
%
\caption{Illustration of relaxation \eqref{pfaNonAutoRegressiveWhitened}.
Components are selected to be well predictable based on the original input signal rather than on themselves.}
\end{figure}
\eqref{pfaNonAutoRegressiveWhitened} is globally solvable by writing it as
\begin{equation} \label{pfaNonAutoRegressiveWhitenedRewrite}
 \optmin{\mathbf{A} \in \orth(n)} \; \tr \left( \mathbf{A}_r^T \bigav{\left(\mathbf{z} - \mathbf{W}\mathbf{\zeta}\right) \left(\mathbf{z} - \mathbf{W}\mathbf{\zeta}\right)^T} \mathbf{A}_r \right)
\end{equation}
and choosing $\mathbf{A}$ such that it diagonalizes
$\bigav{\left(\mathbf{z} - \mathbf{W}\mathbf{\zeta}\right) \left(\mathbf{z} - \mathbf{W}\mathbf{\zeta}\right)^T}$
and sorts the $r$ smallest eigenvalues to the upper left. This can be described as performing PCA on the residuals of the least squares fit. By some calculus, this can be shown to be equal to the method proposed in \cite{boxTiao1977}. To use $\mathbf{A}_r^{(0)}$ with \eqref{pfaAutoRegressiveWhitened}, the prediction model must be refitted to the extracted output by calculating $\mathbf{B}_{\mathbf{z}}(\mathbf{A}_r^{(0)})$ as defined in \eqref{B-ReducedDimRankTransformSolution}.
In \cite{DBLP:conf/icmla/RichthoferW15}, we show that the relaxation gap is related to the absolute prediction error of the optimal solution of \eqref{pfaAutoRegressiveWhitened}. If that error is zero, no relaxation gap exists at all.
If the error is significant, the solution obtained as $\mathbf{A}_r^{(0)}$ can suffer from overfitting being sub-optimal for \eqref{pfaAutoRegressiveWhitened}. In \cite{DBLP:conf/icmla/RichthoferW15}, we propose the following heuristic method to overcome this overfitting.

\subsection{Avoiding overfitting} \label{sec:overfitting}
We propose the heuristics that signals well predictable in terms of \eqref{pfaAutoRegressiveWhitened} yield a lower error-propagation to subsequent predictions than signals that are well predictable in terms of \eqref{pfaNonAutoRegressiveWhitened} but not in terms of \eqref{pfaAutoRegressiveWhitened}. The intuition is that in the second case, prediction is partly based on noisy data -- thus subsequent predictions inherit a higher error.

We define
\begin{equation} \label{V}
\mathbf{V} \quad \coloneq \quad \av{\mathbf{\zeta}(t+1)\mathbf{\zeta}^T(t)}_t \av{\mathbf{\zeta}\mathbf{\zeta}^T}^{-1} \quad \in \; \mathbb{R}^{np \times np}
\end{equation}

$\mathbf{V}$ predicts $\mathbf{\zeta}(t+1)$ from $\mathbf{\zeta}(t)$ like $\mathbf{W}$ predicts $\mathbf{z}(t)$ from $\mathbf{\zeta}(t)$. Note that \eqref{V} is equal to \eqref{W} with $\mathbf{z}(t)$ replaced by $\mathbf{\zeta}(t+1)$. The topmost $n$ entries of $\mathbf{\zeta}(t+1)$ equal those of $\mathbf{z}(t)$ while the others are down-shifted components of $\mathbf{\zeta}(t)$.
Indeed this shift operation is represented in $\mathbf{V}$ as we can write it in terms of $\mathbf{W}$:

\begin{center}
	$\mathbf{V} \quad = \quad $
	\begin{tabular}{ |c|c|c| } 
		\hline
		\multicolumn{2}{|c|}{$\mathbf{W}$} \\
		\hline
		$\begin{matrix} \phantom{\mathbf{I}} & & \\ & \mathbf{I} & \\ & & \phantom{\mathbf{I}} \end{matrix}$ & $\mathbf{0}$ \\
		\hline
	\end{tabular}
\end{center}

The upper part consisting of $\mathbf{W}$ is responsible for predicting $\mathbf{z}(t)$, i.e. the topmost part of $\mathbf{\zeta}(t+1)$ while the lower part performs the shift operation. Now we can perform iterated prediction as follows:
\begin{equation} \label{zVi}
\mathbf{z}(t) \quad \approx \quad \mathbf{W}\mathbf{V}^i \mathbf{\zeta}(t-i)
\end{equation}

\begin{figure}[ht]
   \centering
   \captionsetup{width=.8\linewidth}
   \includegraphics[width=1.0\hsize]{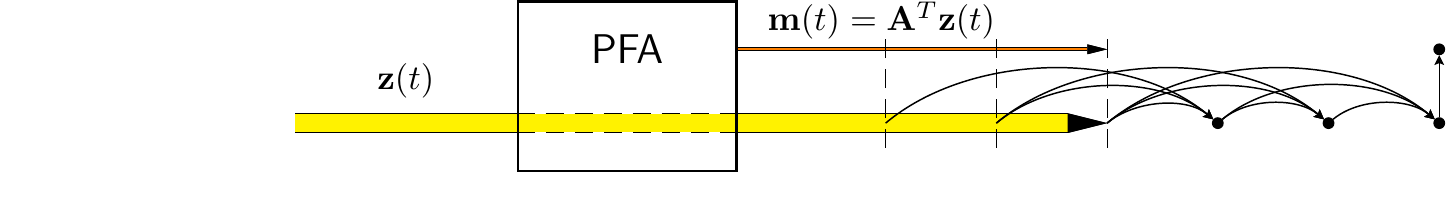}
\caption{Illustration of iterated prediction. Components are selected to be well predictable based on the original input signal and over several iterations of prediction. This filters out false positives, i.e. components that are well predictable based on poorly predictable components. These become expensive as the error would propagate.}
\end{figure}

Taking $k$ steps into account for prediction yields the following optimization problem:
\begin{equation} \label{pfaNonAutoRegressiveIterated}
\optmin{\mathbf{A} \in \orth(n)} \; \sum_{i=0}^k \quad \bigav{\; \norm{\mathbf{A}_r^T(\mathbf{z}-\mathbf{W}\mathbf{V}^i\mathbf{\zeta}(t-i)}^2 \; }_t
\end{equation}
It is globally solvable similarly to \eqref{pfaNonAutoRegressiveWhitened}:
\begin{equation} \label{pfaNonAutoRegressiveIteratedRewrite}
 \optmin{\mathbf{A} \in \orth(n)} \; \tr \Big( \mathbf{A}_r^T \sum_{i=0}^k \bigav{\left(\mathbf{z} - \mathbf{W}\mathbf{V}^i\mathbf{\zeta}(t-i)\right) \left(\mathbf{z} - \mathbf{W}\mathbf{V}^i\mathbf{\zeta}(t-i)\right)^T}_t \mathbf{A}_r \Big)
\end{equation}
Solve \eqref{pfaNonAutoRegressiveIteratedRewrite} by diagonalizing $\sum_{i=0}^k \bigav{\left(\mathbf{z} - \mathbf{W}\mathbf{V}^i\mathbf{\zeta}(t-i)\right) \left(\mathbf{z} - \mathbf{W}\mathbf{V}^i\mathbf{\zeta}(t-i)\right)^T}_t$ and sorting the lowest $r$ eigenvalues to the upper left. Let $\mathbf{A}^{(k)}_r$ denote the global solution of \eqref{pfaNonAutoRegressiveIterated}.
How to optimally choose $k$ for a certain problem is currently an open question, but we know from experiments that in most cases increasing $k$ (up to some unspecified value) improves the prediction error. 
Increasing $k$ too far, however, can lower the quality again, so it is up to experiments how to choose $k$. This effect is illustrated in section \ref{sec:experiments_cells}, figure \ref{fig:cells2}.

\subsection{Taking supplementary information into account for prediction} \label{sec:infoPFA}

The main idea in this work is to enable PFA to take arbitrary supplementary information into account, yielding PFAx. This information shall not be used to extract data from it, but can serve as a helper to improve prediction. Even more important is that PFAx can be used to find subsignals for which the given supplementary information is most relevant. In the following, we let $\mathbf{u}(t)$ denote this supplementary information.

\begin{figure}[ht]
   \centering
   \captionsetup{width=.8\linewidth}
   \includegraphics[width=0.95\hsize]{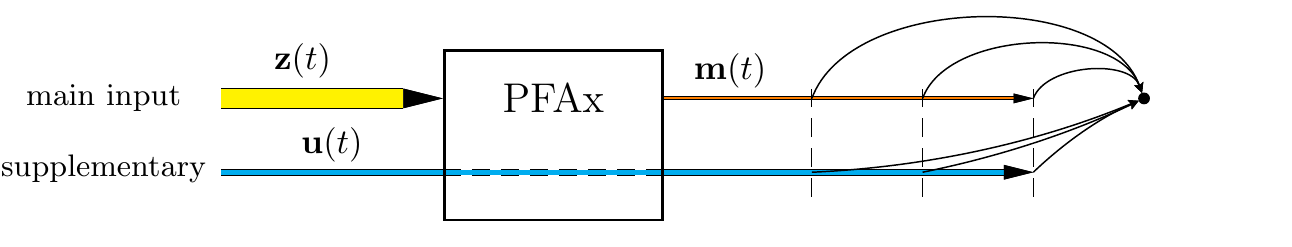}
\caption{Illustration of prediction using supplementary information. Components are selected to be well predictable if supplementary information is taken into account.}
\end{figure}

To perform the desired analysis, we first need to extend the fitting formula \eqref{B-ReducedDimRankTransformSolution} to take supplementary information into account.
Our proposed prediction scheme is as follows:
\begin{align} \label{pfa-predictingWithBU}
 \mathbf{m}(t) \quad \appr^! &\quad \mathbf{B}_{1} \hphantom{\mathbf{U}\mathbf{u}}\!\!\!\!\!\!\! \mathbf{m}(t-1)  + \; \ldots \; + \mathbf{B}_{p} \hphantom{\mathbf{U_q}\mathbf{u}}\!\!\!\!\!\!\!\!\!\! \mathbf{m}(t-p) \qquad \text{with}\quad \mathbf{B}_i \hphantom{\mathbf{U}}\!\!\!\!\! \in \mathbb{R}^{r \times r} \\
                     + &\quad \mathbf{U}_{1} \hphantom{\mathbf{B}\mathbf{m}}\!\!\!\!\!\!\! \mathbf{u}(t-1)  + \; \ldots \; + \mathbf{U}_{q} \hphantom{\mathbf{B_p}\mathbf{m}}\!\!\!\!\!\!\!\!\!\! \mathbf{u}(t-q) \qquad \text{with}\quad \mathbf{U}_i \hphantom{\mathbf{B}}\!\!\!\!\! \in \mathbb{R}^{r \times n_{\mathbf{u}}} \\
                     = &\quad \mathbf{B} \; \underbrace{\mvec(\hist_{\mathbf{z}, p}(t))}_{= \; \mathbf{\zeta}(t)} \;\; + \;\; \mathbf{U} \; \underbrace{\mvec(\hist_{\mathbf{u}, q}(t))}_{\eqcolon \; \mathbf{\mu}(t)}
\end{align}
For a given extraction matrix $\mathbf{A}_r$ the following formulas provide optimal values for $\mathbf{B} \, \in \, \mathbb{R}^{r \times rp}$ and $\mathbf{U} \, \in \, \mathbb{R}^{r \times n_{\mathbf{u}}q}$:
\begin{align} \label{pfa-fittingBU}
 \mathbf{B}(\mathbf{A}_r) \quad \coloneq &\quad \Big( \av{\mathbf{z} \mathbf{\zeta}^T} - \mathbf{U}(\mathbf{A}_r) \av{\mathbf{\mu} \mathbf{\zeta}^T} \Big) \underline{\mathbf{A}_r} \;\; \Big( \underline{\mathbf{A}_r}^T \av{\mathbf{\zeta} \mathbf{\zeta}^T} \underline{\mathbf{A}_r} \Big)^{-1} \\
 \mathbf{U}(\mathbf{A}_r) \quad \coloneq &\quad \Big( \av{\mathbf{z} \mathbf{\mu}^T} - \mathbf{B}(\mathbf{A}_r) \; \underline{\mathbf{A}_r}^T \av{\mathbf{\zeta} \mathbf{\mu}^T} \Big) \; \av{\mathbf{\mu} \mathbf{\mu}^T}^{-1}
\end{align}
For convenience, since $\mathbf{B}(\mathbf{A}_r)$ and $\mathbf{U}(\mathbf{A}_r)$ are only defined implicitly, we provide an explicit formula to obtain $\mathbf{B}(\mathbf{A}_r)$:
\begin{equation} \label{pfa-fittingBUExplicit}
\mathbf{B}(\mathbf{A}_r) \;\; \coloneq \;\; \Big( \av{\mathbf{z} \mathbf{\zeta}^T} - \av{\mathbf{z} \mathbf{\mu}^T} \av{\mathbf{\mu} \mathbf{\mu}^T}^{-1} \av{\mathbf{\mu} \mathbf{\zeta}^T} \Big) \underline{\mathbf{A}_r} \; \Big( \underline{\mathbf{A}_r}^T \left( \av{\mathbf{\zeta} \mathbf{\zeta}^T} - \av{\mathbf{\zeta} \mathbf{\mu}^T} \av{\mathbf{\mu} \mathbf{\mu}^T}^{-1} \av{\mathbf{\mu} \mathbf{\zeta}^T} \right) \underline{\mathbf{A}_r} \Big)^{-1}
\end{equation}
If the arising inversions are not computable due to near-zero-eigenvalues, proceed like explained in our comment on inverting $\av{\mathbf{\zeta} \mathbf{\zeta}^T}$.
In this section we define $\mathbf{W} \; \coloneq \; \mathbf{B}(\mathbf{I})$ in analogy to \eqref{W}, but based on \eqref{pfa-fittingBUExplicit} rather than \eqref{B-ReducedDimRankTransformSolution}. Now we can adopt \eqref{pfaNonAutoRegressiveWhitenedRewrite} to find an optimal $\mathbf{A}_r$ in terms of the relaxation:
\begin{equation} \label{ipfaNonAutoRegressiveWhitenedRewrite}
 \optmin{\mathbf{A} \in \orth(n)} \; \tr \left( \mathbf{A}_r^T \bigav{\left(\mathbf{z} - \mathbf{W}\mathbf{\zeta} - \mathbf{U}(\mathbf{I})\mathbf{\mu}\right) \left(\mathbf{z} - \mathbf{W}\mathbf{\zeta} - \mathbf{U}(\mathbf{I})\mathbf{\mu}\right)^T} \mathbf{A}_r \right)
\end{equation}
Just like \eqref{pfaNonAutoRegressiveWhitenedRewrite}, \eqref{ipfaNonAutoRegressiveWhitenedRewrite} can be solved by diagonalizing the central matrix and sorting the smallest eigenvalues to the upper left. Applying the technique from section \ref{sec:overfitting} to the optimization problem of this section is a bit involved. We define
\begin{align}
\mathbf{V} \hphantom{(t)}\;\;\, \quad \coloneq &\quad \Big( \av{\mathbf{\zeta}(t+1) \mathbf{\zeta}^T} - \mathbf{I}_{np,n}\mathbf{U}(\mathbf{I}) \av{\mathbf{\mu} \mathbf{\zeta}^T} \Big) \av{\mathbf{\zeta} \mathbf{\zeta}^T}^{-1} \\
\hat{\mathbf{z}}^{(i)}(t) \quad \coloneq &\quad \mathbf{W}\mathbf{V}^i \mathbf{\zeta}(t-i) + \mathbf{I}^T_{np, n} \sum^i_{j=0} \mathbf{V}^j \mathbf{I}_{np,n} \mathbf{U}(\mathbf{I}) \mathbf{\mu}(t-j)
\end{align}
Note that $\mathbf{W}$ and $\mathbf{V}$ simplify to their earlier definitions for $\mathbf{U} = 0$. Also the predictor $\hat{\mathbf{z}}^{(i)}$ simplifies to the predictor defined in \eqref{zVi} for $\mathbf{U} = 0$. For $i=0$ it even boils down to $\mathbf{W}\mathbf{\zeta}$, which was used as predictor in \eqref{pfaNonAutoRegressiveWhitened}. Consequently, we can also write the optimization problem in terms of $\hat{\mathbf{z}}^{(i)}$:
\begin{equation} \label{ipfaNonAutoRegressiveWhitenedRewrite}
 \optmin{\mathbf{A} \in \orth(n)} \sum_{i=0}^k \bigav{\; \norm{\mathbf{A}_r^T(\mathbf{z}-\hat{\mathbf{z}}^{(i)})}^2 \; } \quad = \quad \tr \Bigg( \mathbf{A}_r^T \sum_{i=0}^k \bigav{\big(\mathbf{z} - \hat{\mathbf{z}}^{(i)}\big) \big(\mathbf{z} - \hat{\mathbf{z}}^{(i)}\big)^T} \mathbf{A}_r \Bigg)
\end{equation}
It can be solved by the familiar procedure:
Choose $\mathbf{A}$ such that it diagonalizes $\sum_{i=0}^k \bigav{\left(\mathbf{z} - \hat{\mathbf{z}}^{(i)}\right) \left(\mathbf{z} - \hat{\mathbf{z}}^{(i)}\right)^T}$ and sort the lowest $r$ eigenvalues to the upper left.

\section{Using extracted features to perform control} \label{sec:controlPFA}

The idea here is that $\mathbf{z}$ represents some kind of an agent's perception, while the supplementary information $\mathbf{u}$ represents the agent's control commands. The extracted predictable features are a compact representation of perception aspects that are influenced by the control commands. Now we consider the setting that after the training phase $\mathbf{A}_r$, $\mathbf{B}(\mathbf{A}_r)$ and $\mathbf{U}(\mathbf{A}_r)$ are given and we want to reach a goal position $\mathbf{m}^*$ in feature space. We want to calculate what control must be chosen to reach a desired position in feature space. For this purpose, we minimize the least square distance between predicted features and goal features:

\begin{align} \label{ipfaOptControl}
 \optmin{\mathbf{u}(t) \; \in \; \mathbb{R}^{n_{\mathbf{u}}}} &\norm{\mathbf{m}^* - \mathbf{B}(\mathbf{A}_r) \mathbf{A}_r^T \mathbf{\zeta}(t+1) - \mathbf{U}(\mathbf{A}_r) \mathbf{\mu}(t+1)}^2 \\
= \quad &\norm{\underbrace{\mathbf{m}^* - \mathbf{B}(\mathbf{A}_r) \mathbf{A}_r^T \mathbf{\zeta}(t+1) - \Big(\sum_{j=2}^q \mathbf{U}_j(\mathbf{A}_r) \mathbf{u}(t-j+1)\Big)}_{\eqcolon \; \mathbf{u}^* \; \in \; \mathbb{R}^{r}} - \mathbf{U}_1 \mathbf{u}(t)}^2 \\
 = \quad &\norm{\mathbf{u}^*  - \underbrace{\mathbf{U}_1}_{r \times n_{\mathbf{u}}} \mathbf{u}(t)}^2
\end{align}
Solving this problem is straight forward by choosing $\mathbf{u}(t) \coloneq \mathbf{U}_1^{-1} \mathbf{u}^*$ (or $\mathbf{u}(t)~\coloneq~(\mathbf{U}_1^T\mathbf{U}_1)^{-1} \mathbf{U}_1^T \mathbf{u}^*$, if $\mathbf{U}_1$ is not quadratic or not invertible).
Note that for this solution squaring would not have been necessary, i.e.\ it would also minimize $\norm{\mathbf{u}^*  - \mathbf{U}_1 \mathbf{u}(t)}$. However, as soon as there are constraints on $\mathbf{u}$, the squared distance is much friendlier for optimization.

\begin{figure}[ht]
   \centering
   \captionsetup{width=.8\linewidth}

   \includegraphics[width=0.9\hsize]{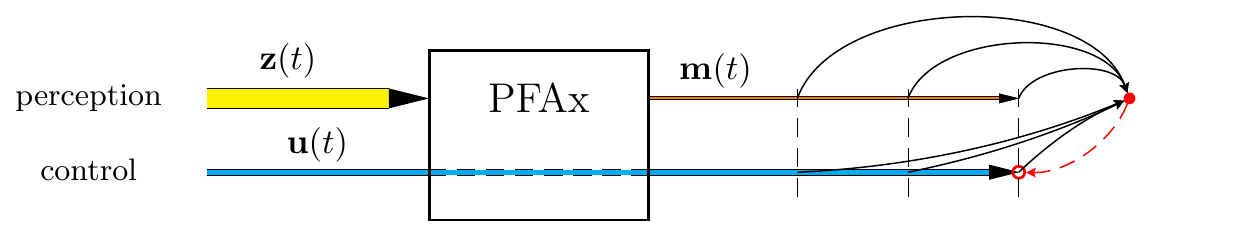}

\caption{Illustration of controlling predictable features. The influence of the control signal onto prediction is inverted to calculate a control command that would likely yield a desired outcome.}
\end{figure}

We will usually have constraints on $\mathbf{u}(t)$, reflecting the agent's limited performance. A typical case for future applications are linear constraints, which would yield an efficiently solvable quadratic optimization problem (QP). In the following sections we will model a robot driving with constant speed, so we need to deal with a normalized-length-constraint:
\begin{equation}
\optmin{\substack{\mathbf{u}(t) \; \in \; \mathbb{R}^{n_{\mathbf{u}}}\\ \norm{\mathbf{u}(t)} \; = \; c}} \norm{\mathbf{u}^*  - \mathbf{U}_1 \mathbf{u}(t)}^2
\end{equation}
This problem is equivalent to the inhomogeneous eigenvalue problem
\begin{align} \label{ipfaOptControlInhomEVP}
\mathbf{U}_1^T \mathbf{U}_1 \mathbf{u}(t) \quad = \quad &\lambda \mathbf{u}(t) + \mathbf{U}_1^T \mathbf{u}^* \\
\norm{\mathbf{u}(t)} \quad = \quad &c
\end{align}
In \cite{MATTHEIJ1987507} such problems are approached. For convenience, we provide a method from there in \ref{sec:inhomEVP}.

\section{Experiments and Applications}
In this section we apply the principles developed so far to some simulated problem settings.
We show how this is suitable to generate motor control for letting an agent reach a certain goal position in a simple environment, based on different sensors. Once the environment gets complexer it turns out that the principle is mainly suitable for local navigation and we show experiments that illustrate this limitation. In a follow-up paper we will show how this principle can be extended to perform global navigation reliably.

\subsection{Using place cells for navigation} \label{sec:experiments_cells}
In this section we apply the developed techniques to the following setting:
We have a virtual rat -- an agent -- on a table and assume there is already a mechanism that obtains place cells from vision. This is a reasonable assumption, because in \cite{FranziusSprekelerEtAl-2007e, FranziusSprekelerEtAl-2007d, Schoenfeld2015} SFA has been shown to be capable of such a preprocessing.
To find out whether it is possible to use PFAx for goal-driven navigation, we place a certain number of 2D-Gaussians on the table to model place cells. That means, each component of $\mathbf{x}(t)$ is the activation of one place cell, i.e.\ a 2D-Gaussian centered on a random position, evaluated for the rat's position at time $t$. In a training phase consisting of a random walk the rat explores the environment and PFAx extracts the most predictable signals from place cell input $\mathbf{x}$ ($\mathbf{z}$ after sphering). Within this section we always use $50$ place cells, i.e.\ $dim(\mathbf{x}) = 50$. 
To constrain the complexity of the experiment we let the rat walk with constant speed. So the only control information is direction. This is provided as a direction vector and fed into PFAx as supplementary information $\mathbf{u}$
(i.e.\ the movement deltas are provided to PFAx).

After the training phase we choose a goal position and use the extraction matrix obtained by PFAx to calculate corresponding goal features. Then we iterate the optimization from section \ref{sec:controlPFA}, where $c$ is the rat's constant speed. We actually apply the optimized $\mathbf{u}(t)$-command after each step, and increase $t$. Finally we observe the resulting path and modify the setup in various ways to measure the robustness of this method.

Based on a random walk, PFAx can only find predictability w.r.t.\ to the provided movement delta information. Since this incorporates just one time step, $p = 1$ is an exhaustively sufficient value for all experiments. A higher value for $p$ would only make sense if the movement simulation of the agent would incorporate higher moments like acceleration. Adding such higher moments to the simulation is postponed to future work.

\begin{figure}[hb!]
	\centering
	\includegraphics[width=0.7\hsize]{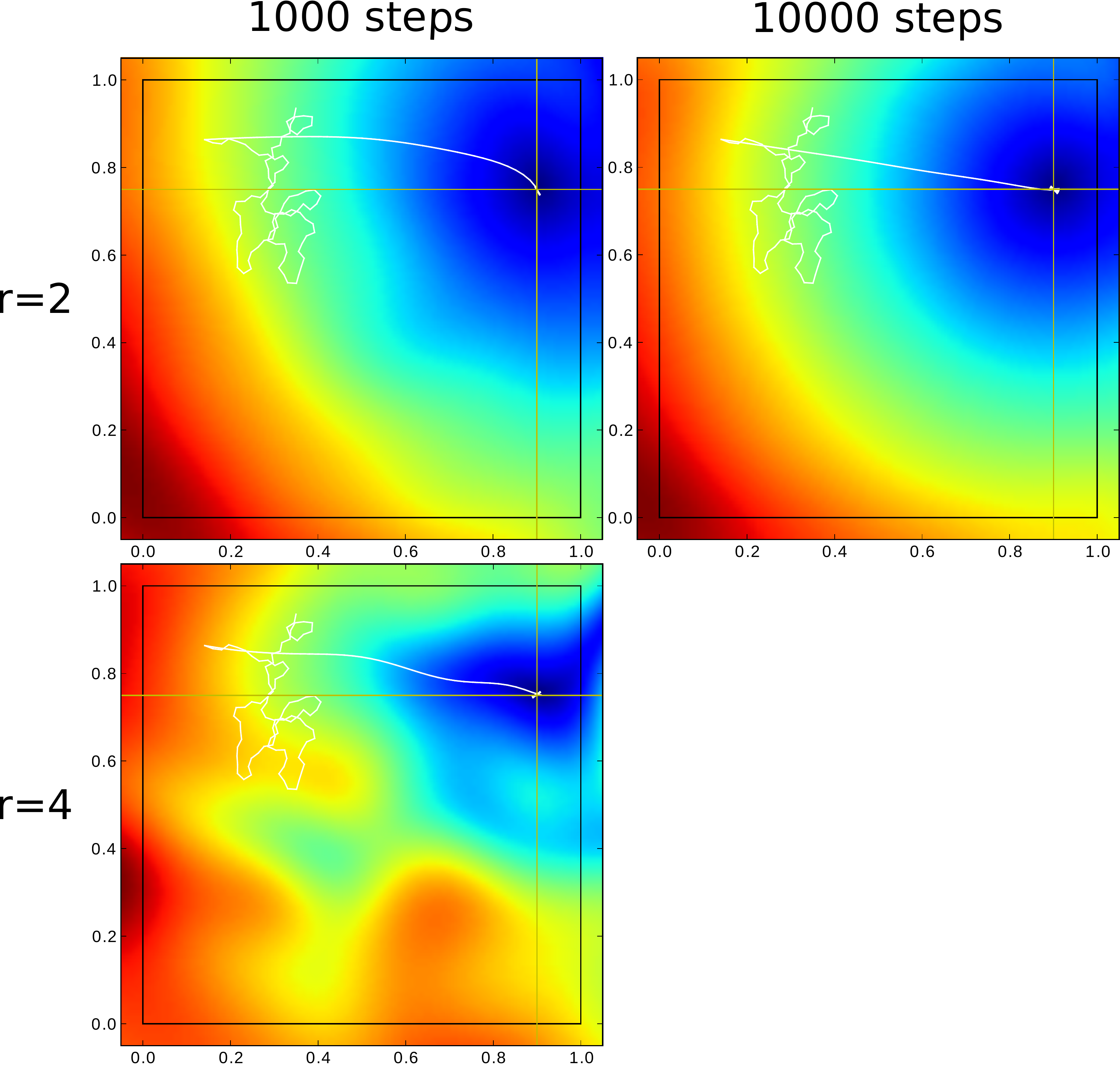}
	\caption{Compare $r \in \{2, 4\}$, $\abs{\trph} \in \{1000, 10000\}$, $50$ place cells, $k=0$}
	\label{fig:cells1}
\end{figure}

\begin{figure}[hb!]
	\centering
	\includegraphics[width=1.0\hsize]{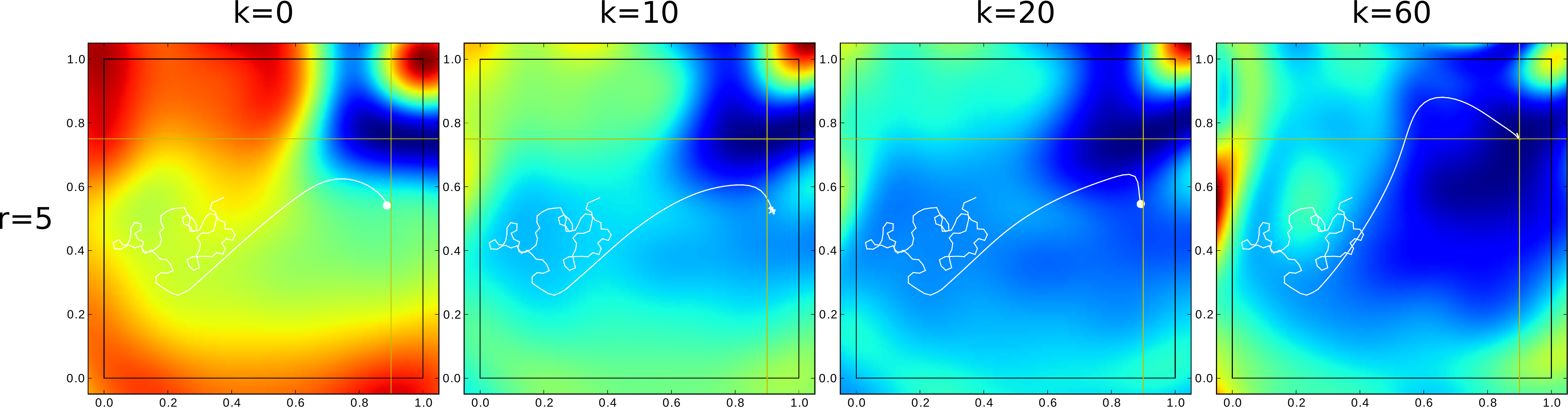}
	\caption{Varying $k$, $\abs{\trph} = 8000$, $r=5$, $50$ place cells}
	\label{fig:cells2}
\end{figure}

\begin{figure}[h!]
	\centering
	\includegraphics[width=1.0\hsize]{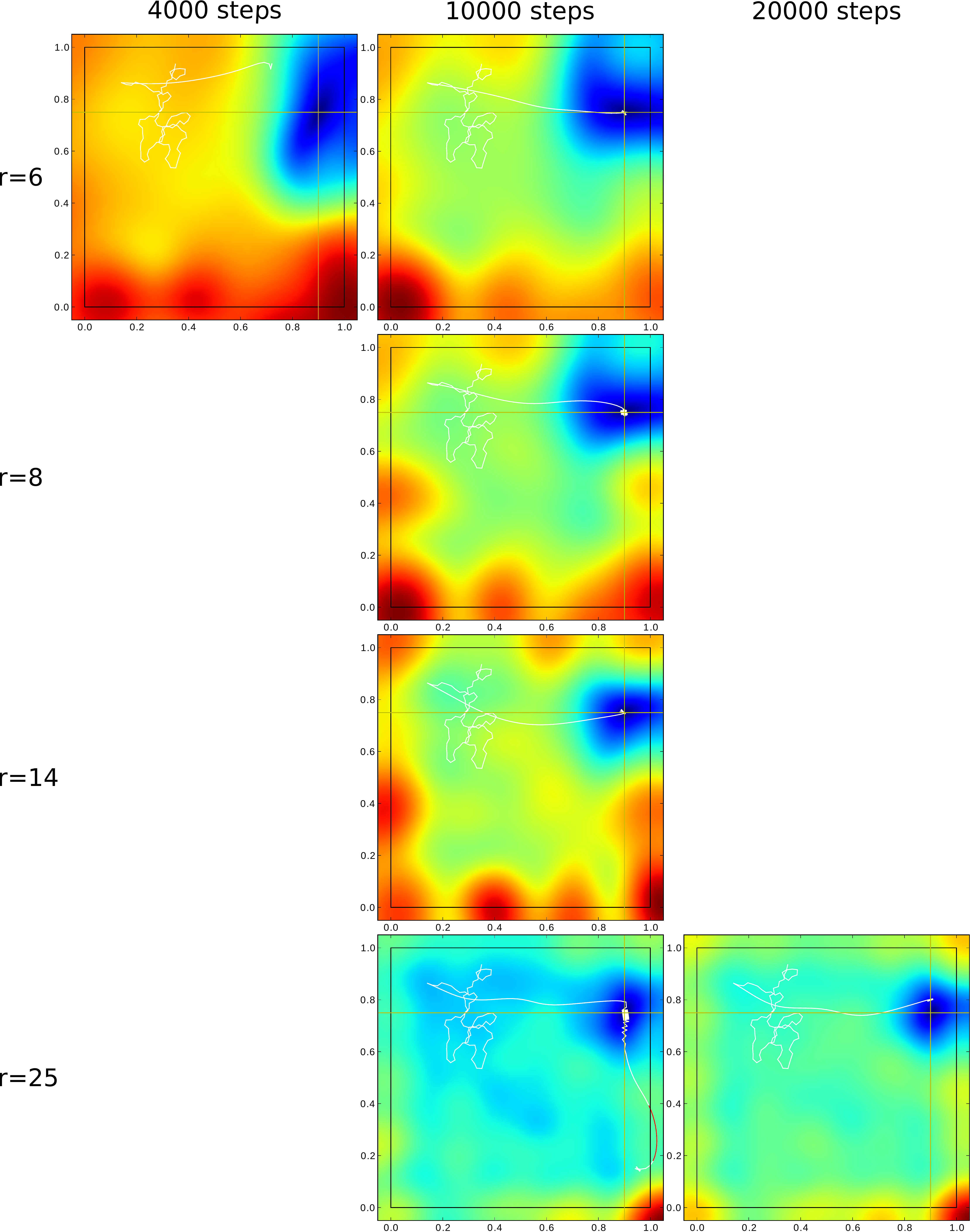}
	\caption{More values of $r$ and $\abs{\trph}$, $50$ place cells, $k=0$}
	\label{fig:cells3}
\end{figure}

Our first set of experiments is shown in figure \ref{fig:cells1}. The yellow crosshairs always indicate the goal and the colormap displays Euclidean distance of each point to the goal in $r$-dimensional feature space. The navigation path is rendered in white on top of the colormap. To indicate the starting point, some final steps of the random walk training phase are shown and the algorithmic navigation is clearly distinguishable from the random walk by yielding a straight or slightly curved path. We conclude from figure \ref{fig:cells1} that this kind of navigation is robust for a value of $r$ that properly matches the problem complexity, in this setting ideally $r=2$. Larger values of $r$ add irrelevant information to the feature space and thus distract the navigation algorithm. However, for moderate values of $r$ this can be compensated by increasing the value of $k$ -- like shown in figure \ref{fig:cells2} or by increasing the training phase -- like shown in figure \ref{fig:cells3}. Detecting a good value for $r$ is subject of ongoing research.

\subsection{Place-cell-based navigation with obstacle}

\begin{figure}[h!]
	\centering
	\includegraphics[width=0.4\hsize]{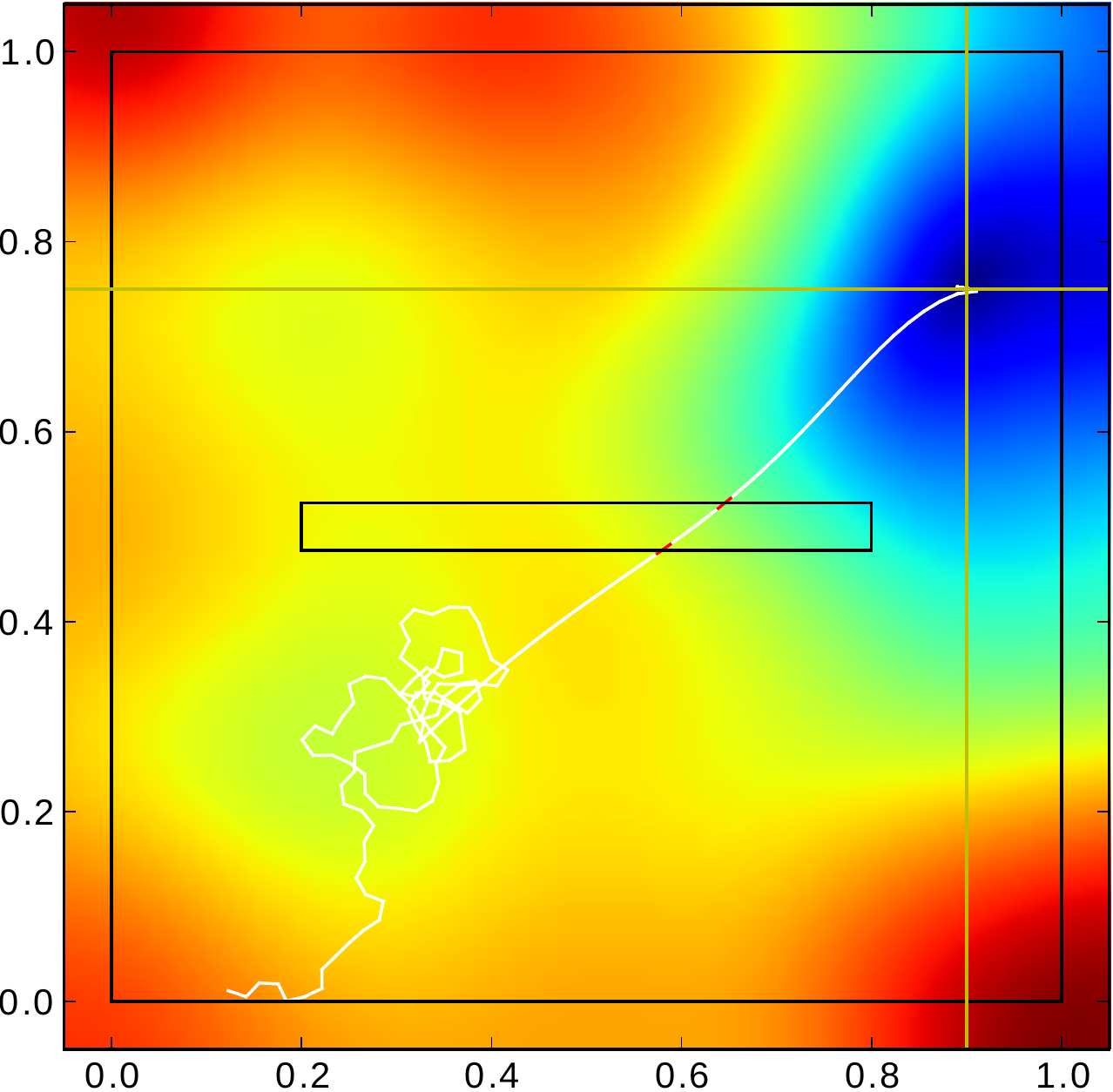}
	\caption{$r=3$, $\abs{\trph} = 8000$, $50$ place cells, $k=0$, collision detection turned off}
	\label{fig:cell-obstacle}
\end{figure}

To examine the capabilities of global navigation we add an obstacle to the environment. It turns out that there exists no parameter set that would enable the Gaussian-modeled place-cell-based navigation routine to measure the obstacle. Examining the color inside the obstacle we see that it is not represented in feature space at all.
Figure \ref{fig:cell-obstacle} illustrates an exemplary run that demonstrates how the algorithm would want to go through the obstacle if collision detection is turned off. Note that during training phase collision with the obstacle is prohibited.

We conclude that naively using Gaussians to model place cells is inherently unsuitable for measuring obstacles, because in this model the place cells just shine through the walls of the obstacle. So we refine the simulation to use a sensor that is sensitive to obstacle walls and cannot look through them. Constructing place cells that do not shine through walls would not be feasible with plain Gaussians. Raw vision input would lead to a high dimensional input, but we prefer to keep the setting low dimensional for now and focus on the navigation task. We therefore propose a sensor that is invariant under head direction, given that SFA has been shown to be capable of finding head direction invariant features from vision input, see \cite{FranziusSprekelerEtAl-2007e, FranziusSprekelerEtAl-2007d, Schoenfeld2015}. So, with SFA as a potential preprocessing step in mind, these are reasonable simplifications.

\subsection{Using a wall sensor for navigation}

\begin{figure}[h!]
	\centering
	\includegraphics[width=0.6\hsize]{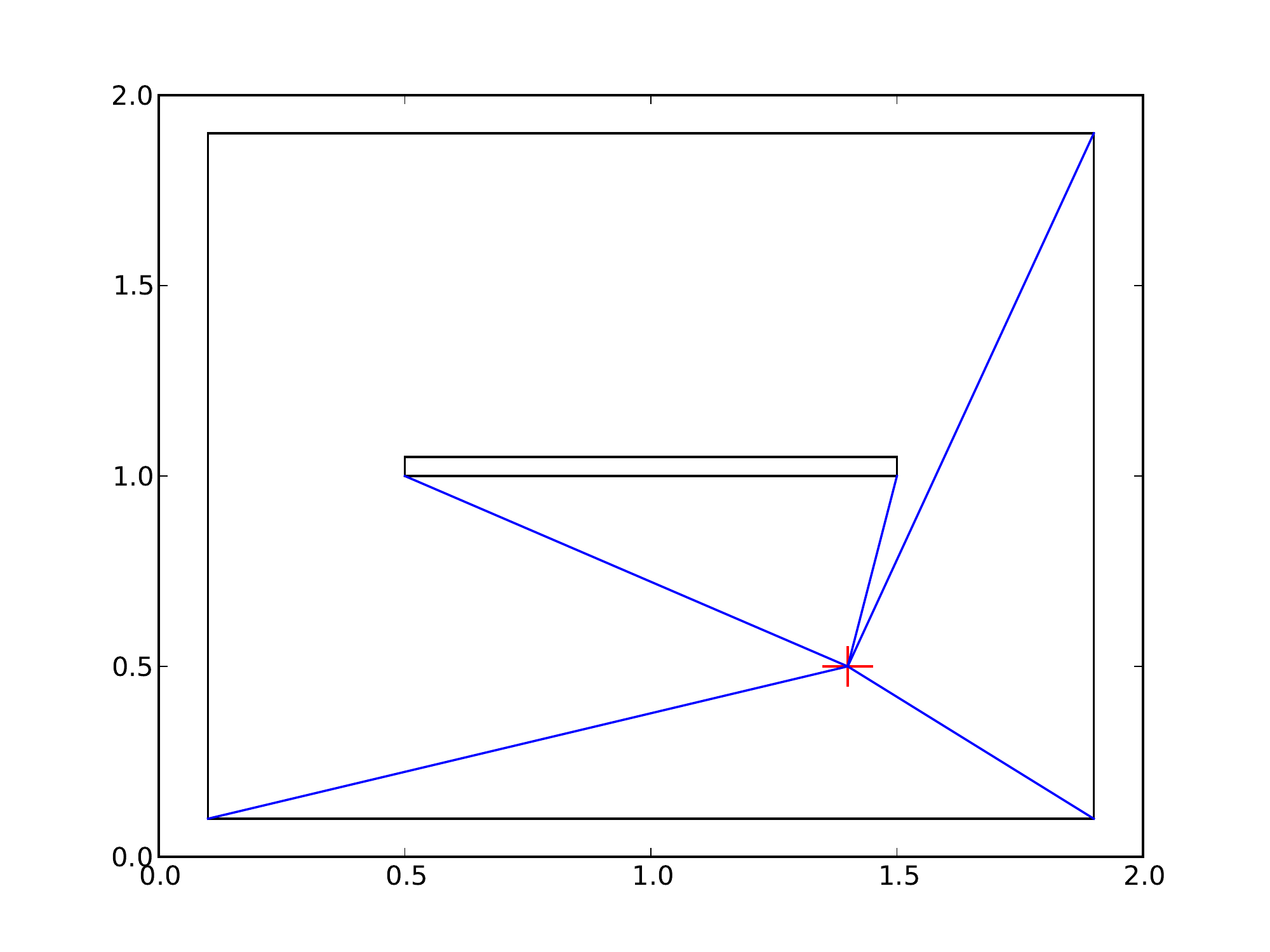}
	\caption{Sensor that measures the fraction each wall segment occupies in an $360$° field of view}
	\label{fig:wall-sensor}
\end{figure}

We introduce a virtual sensor that measures -- for its current location -- the visible fraction of each wall segment in an overall $360$° field of view. Figure \ref{fig:wall-sensor} shows how the full field of view is split up into sections occupied by each wall segment. In the example the sensor consists of $8$ components, one for each wall segment existing in the environment. The components belonging to the left and right walls of the obstacle are invisible from the location illustrated in figure \ref{fig:wall-sensor} and would emit a value of $0$ in that case.

\begin{figure}[h!]
	\centering
	\includegraphics[width=0.4\hsize]{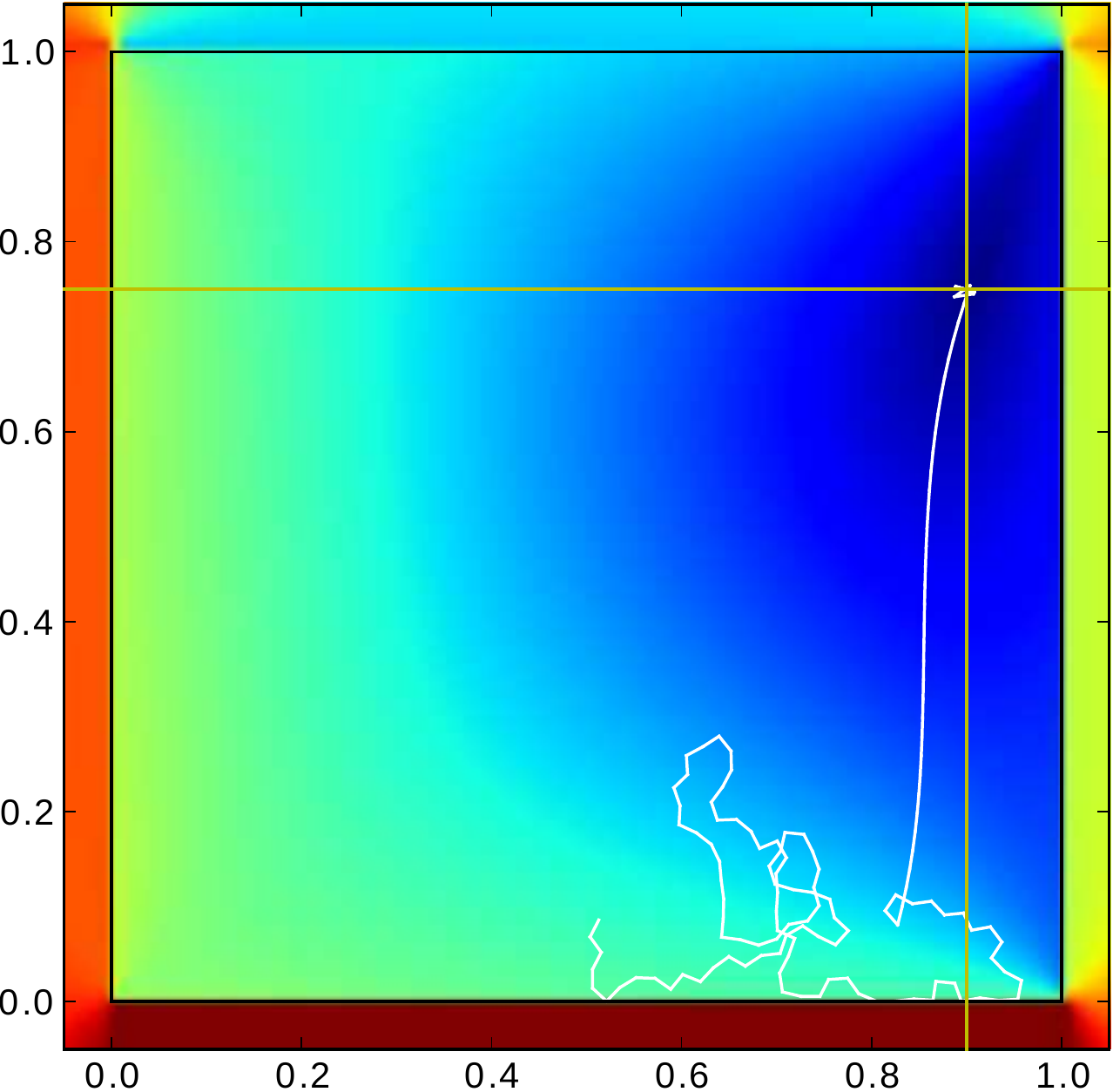}
	\caption{Using the wall sensor for navigation without obstacle}
	\label{fig:wall-sensor-no-obstacle}
\end{figure}

Using a $360$° field of view the sensor is by construction invariant under head direction. As an initial experiment we show that navigation without obstacle is feasible. With $4$ surrounding wall segments the environment in figure \ref{fig:wall-sensor-no-obstacle} yields $dim(x) = 4$. Contrary to the place cells scenario we apply a quadratic expansion on the sensor to compensate its low dimensionality. Expansions with higher than quadratic degree provided no significant improvement for this setting. So we focus on quadratic expansion for all wall sensor experiments in this paper. Further, we use a fixed-length, exhaustive training phase of $\abs{\trph}=40000$ steps for all such experiments.

Given that the navigation without obstacle works reliably with all sorts of parameter sets, we do not provide experiments beyond the exemplary illustration in figure \ref{fig:wall-sensor-no-obstacle} and instead focus on the setting with obstacle.

\subsection{Wall-sensor-based navigation with obstacle}

Using the wall sensor it turns out that an obstacle is now clearly represented in feature space -- note the different color inside the obstacle. Depending on $r$ and $k$ the feature space surrounding the obstacle can guide the agent around. To show that the feature space itself is repelling the agent from the obstacle, navigation is performed without collision detection. During training phase, however, collision with any walls is prohibited.

\begin{figure}[h!]
	\centering
	\includegraphics[width=0.8\hsize]{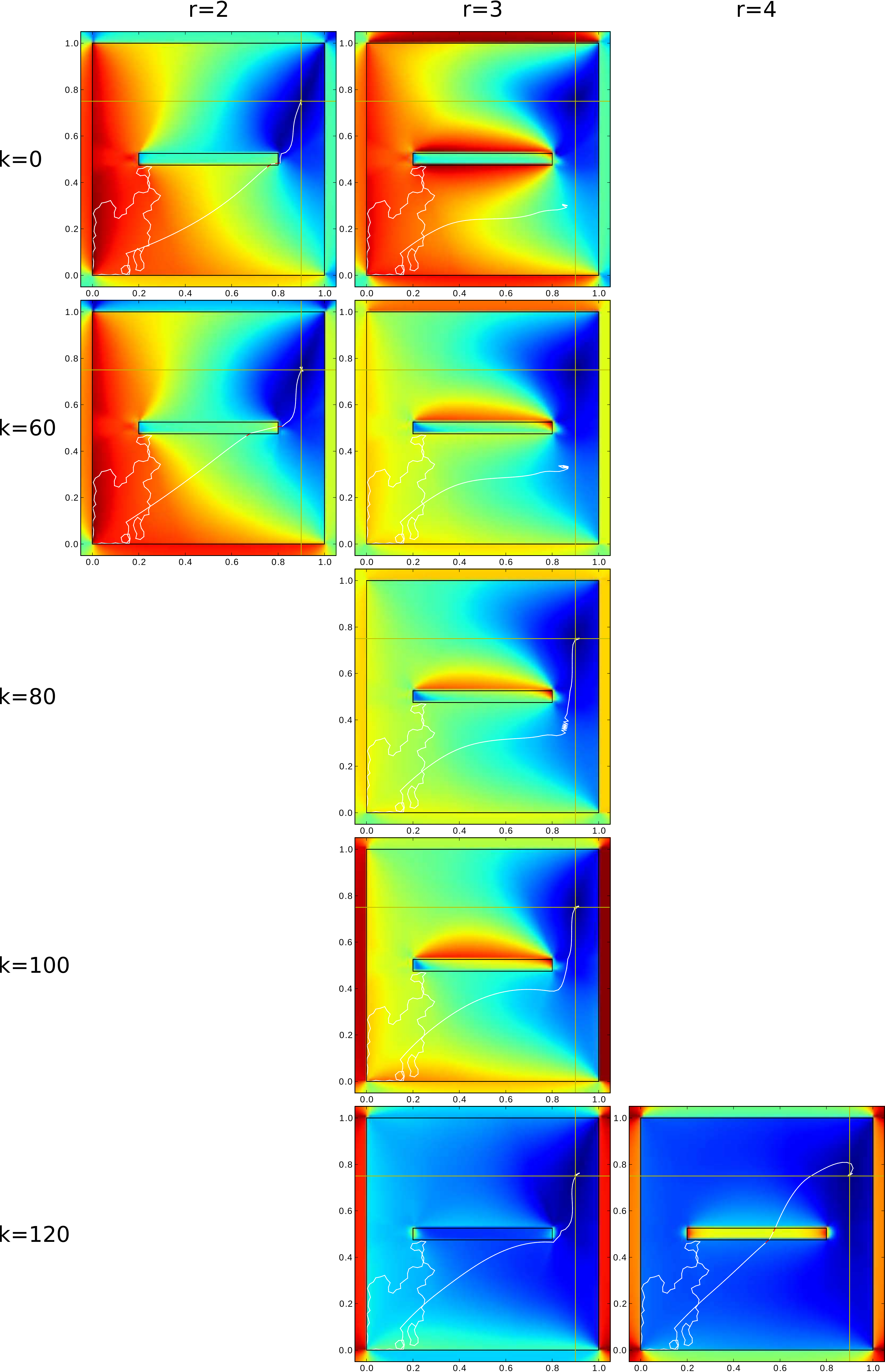}
	\caption{Using the wall sensor for navigation with far obstacle}
	\label{fig:wall-sensor-obstacle-far}
\end{figure}

In figure \ref{fig:wall-sensor-obstacle-far} we show a setting with a starting point rather far away from the obstacle.
It suggests that with the right parameters it is actually possible to avoid the obstacle.

With a starting point near the obstacle, like presented in figure \ref{fig:wall-sensor-obstacle-near}, we observe somewhat equal results -- with the right values for $r$ and $k$ a proper navigation can be achieved, but the choice is not entirely stable. While a higher value of $k$ mostly improves the navigation, choosing it too high can break it.

\begin{figure}[h!]
	\centering
	\includegraphics[width=0.65\hsize]{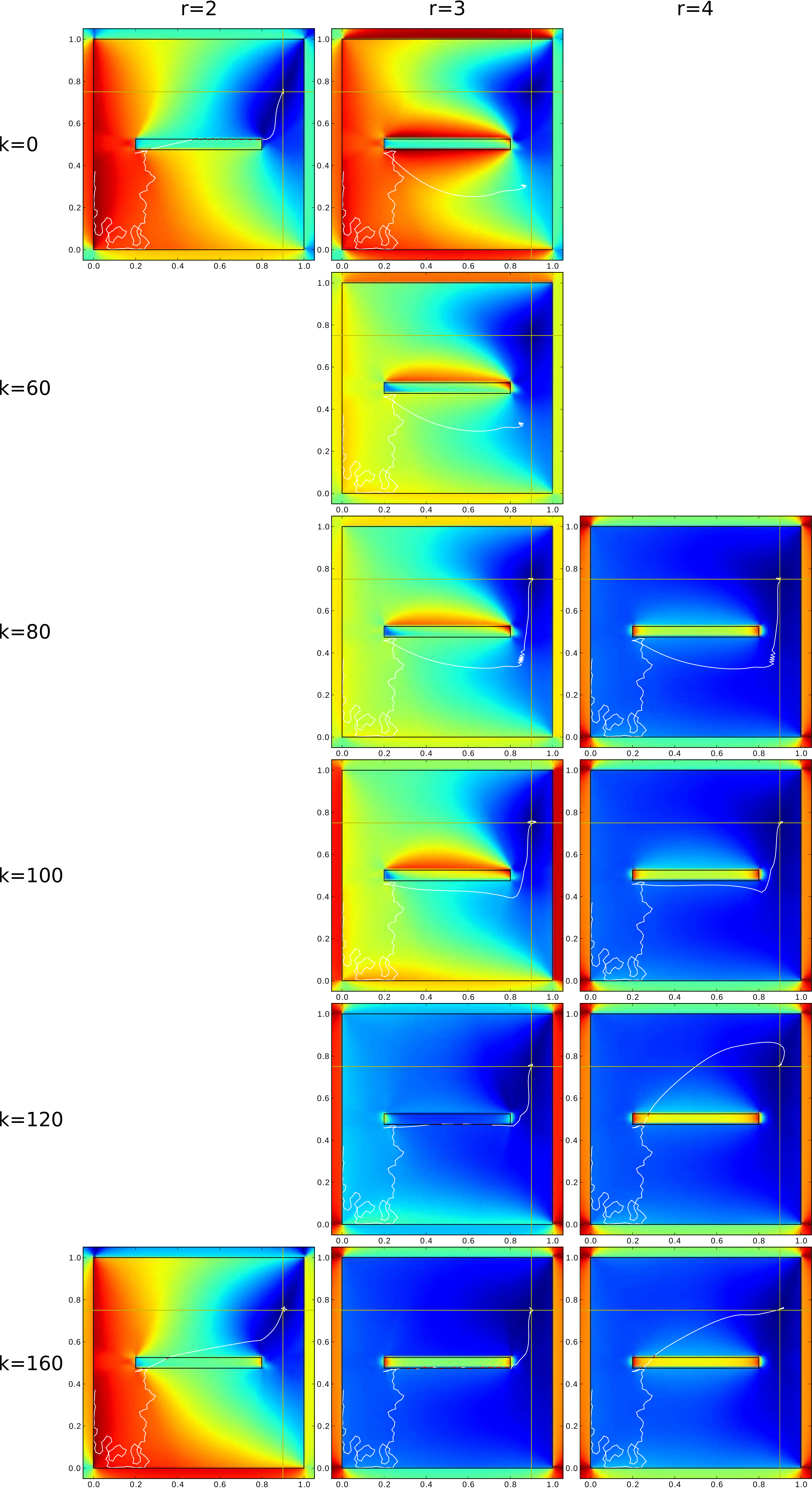}
	\caption{Using the wall sensor for navigation with near obstacle}
	\label{fig:wall-sensor-obstacle-near}
\end{figure}

\subsubsection*{}
Finally with a goal position closely behind the obstacle we present an example where the agent is repelled from the obstacle, but can hardly find its way around it (figure \ref{fig:wall-sensor-obstacle-near}). The sufficient value $k=200$ is rather unstable and already with $k=220$ the agent would pass through the obstacle. Running with collision detection, the wall itself would still guide the agent around the obstacle, but even then the navigation is not very stable.

\begin{figure}[h!]
	\centering
	\includegraphics[width=0.7\hsize]{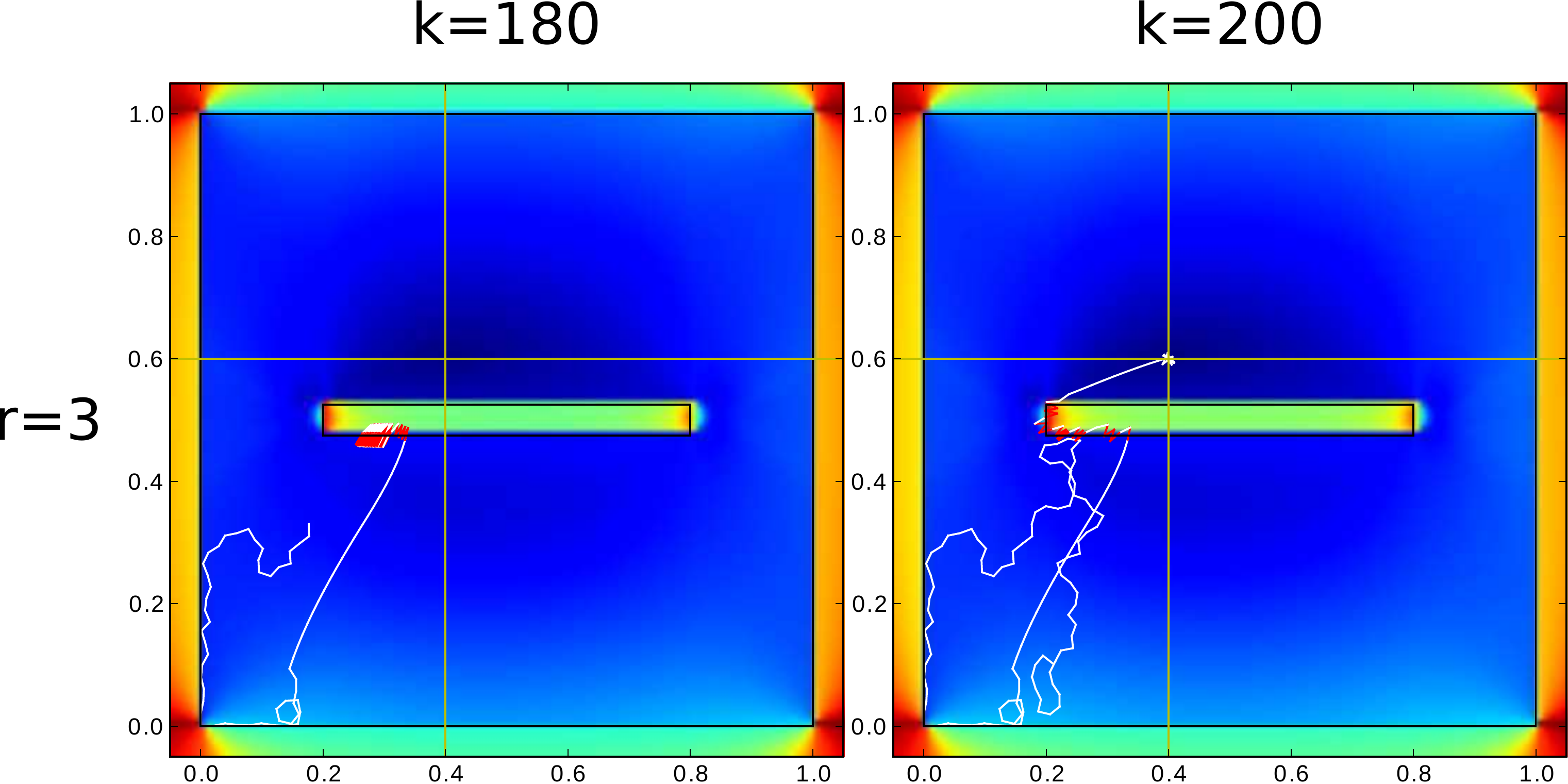}
	\caption{Using wall sensor for navigation with a goal position closely behind the obstacle}
	\label{fig:wall-sensor-obstacle-near}
\end{figure}

\section{Conclusion}
We conclude that the presented approach can be used to learn the relation between a control command signal and an agent's state in an environment. Especially it can find a low dimensional feature space that can efficiently represent the agent's state in a way that allows goal driven manipulation. We can generate the proper control signal to optimize the agent's state locally and approximately perform a gradient descent within feature space. This is, however, hardly sufficient to perform global navigation, e.g.\ navigate around an obstacle. While some promising results suggest that even global optimization is feasible, they highly depend on the choice of parameters $r$ and $k$. A theory for handling these parameters more systematically is subject to current research. We are preparing a follow-up paper with an extension to this algorithm that enables stable global navigation.

\section*{Acknowledgments}
This work has been funded by a grant from the German Research Foundation (Deutsche Forschungsgemeinschaft, DFG) to L. Wiskott (SFB 874, TP B3) and
supported by the German Federal Ministry of
Education and Research within the National Network Computational
Neuroscience -- Bernstein Fokus: “Learning behavioral models: From human
experiment to technical assistance”, grant FKZ 01GQ0951.

\pagebreak
\bibliography{MachineLearning,Miscellaneous,Predictability,ReinforcementLearning,SFA,Extern} 

\pagebreak
\appendix
\section{Appendix}
\subsection{Notation overview} \label{sec:notation}
This section gives an overview of the notation used in this paper.

\begin{tabular}{p{1.1cm}p{9.7cm}ll}
$\mathbf{x}(t)$ & denotes the raw input signal.\\

$\mathbf{u}(t)$ & denotes the supplementary information signal.\\

$\trph$ & $\coloneq~\{t_0, \ldots, t_k\}$ denotes a discrete time sequence (considered as equidistant with step size normalized to $1$). We usually
refer to $\trph$ as the \textit{training phase}.\\

$\av{\mathbf{s}(t)}_{t \in S}$ & $\coloneq~\frac{1}{\abs{S}} \sum_{t\in S} \mathbf{s}(t)$ denotes the average of some signal $\mathbf{s}$ over a finite set $S$.
For $S~=~\trph$ we just write $\av{\mathbf{s}(t)}_t$ or even $\av{\mathbf{s}}$, if it is obvious, what unbound variable is targeted.\\

$\mathbf{h}(\mathbf{x})$ & denotes the expansion function and usually consists of a set of monomials of low degree.\\

$\mathbf{z}(t)$ & denotes $\mathbf{h}(\mathbf{x}(t))$ after sphering.\\


$\mathbf{m}(t)$ & denotes the optimized output signal ($\mathbf{m}$ for \textit{model}).\\

$n$ & $\coloneq~dim(\mathbf{h}(\mathbf{x}))$
denotes the number of components to be analyzed (after expansion).\\

$n_{\mathbf{u}}$ & $\coloneq~dim(\mathbf{u})$
denotes the number of components in $\mathbf{u}$\\

$r$ & denotes the number of extracted components (“features”).\\

$\mathbf{A}, \mathbf{a}$ & denotes the matrix (or vector if $r=1$) holding the linear composition of the output-signal.
We set $\mathbf{m}(t)~=~\mathbf{A}^T\mathbf{z}(t)$.\\

$\mathbf{a}_i$ & denotes the $i$'th column of $\mathbf{A}$, so we can write $m_i(t)~=~\mathbf{a}_i^T \mathbf{z}(t)$.\\

$\orth(n)$ & $\subset \mathbb{R}^{n \times n}$ denotes the orthogonal group of dimension $n$, i.e.\ $\forall \; \mathbf{A} \in \orth(n) \colon \; \mathbf{A}\mathbf{A}^T = \mathbf{A}^T\mathbf{A} = \mathbf{I}$\\

$p$ & denotes the number of recent signal-values involved in the prediction. We also call it the \textit{prediction-order}.\\

$\mathbf{I}_{s, r}$ & denotes the $s\times r$ identity matrix ($s$ counting rows, $r$ counting columns).
For $s = r$ this is a usual square identity, while in the non-square case it consists
of a square identity block in the top or left area, filled up with zeroes to fit the given shape.\\

$\mathbf{I}_{r}$ & $\coloneq~\mathbf{I}_{n, r}$\\


$\mathbf{A}_r$ & $\coloneq~\mathbf{A}\mathbf{I}_r$


%
\end{tabular}
\renewcommand{\arraystretch}{1}

We frequently use the $p$-step time-history of a signal $\mathbf{z}$, which we formalize by the following function:
\begin{align} \label{history}
	&\hist_{\mathbf{z}, p, \Delta}(t) \quad \coloneq \quad \sum_{i=1}^p \quad \mathbf{z}(t-i\Delta) \mathbf{e}_i^T \quad \text{with} \quad \mathbf{e}_i \in \mathbb{R}^p\\
	&\hist_{\mathbf{z}, p}(t)_{\hphantom{, \Delta}} \quad \coloneq \quad \hist_{\mathbf{z}, p, 1}(t)
\end{align}
Here $\mathbf{e}_i$ denotes the $i$-th $p$-dimensional euclidean unit vector, which is $1$ at position $i$ and $0$ everywhere else.

Further more we sometimes use the Kronecker product $\otimes$ and the $\mvec$-operator defined as follows:

For matrices $\mathbf{A} \in \mathbb{R}^{m \times n}$ and $\mathbf{B} \in \mathbb{R}^{k \times l}$ and with $a_{ij}$ denoting the entries, $\mathbf{a}_i$ the columns of $\mathbf{A}$:

\begin{equation}
 \mathbf{A} \otimes \mathbf{B} \quad \coloneq \quad \left( \begin{matrix} a_{11}\mathbf{B}& \cdots & a_{1n}\mathbf{B} \\ \vdots & \ddots & \vdots \\ a_{m1} \mathbf{B} & \cdots & a_{mn}\mathbf{B} \end{matrix} \right) \; \in \; \mathbb{R}^{mk \times nl}
\end{equation}

\begin{equation}
\mvec(\mathbf{A}) \quad \coloneq \quad \left( \begin{matrix} \mathbf{a}_1\\ \vdots \\ \mathbf{a}_n \end{matrix} \right) \; \in \; \mathbb{R}^{mn}
\end{equation}

Additionally, we sometimes make use of the following shortcut:
\begin{equation} \label{multiA}
\underline{\mathbf{A}} \quad \coloneq \quad \mathbf{I}_{p, p} \otimes \mathbf{A} \quad = \qquad \underbrace{\!\!\!\!\!\!\left( \begin{matrix} \mathbf{A}&  & \mathbf{0} \\  & \ddots &  \\ \mathbf{0} &  & \mathbf{A} \end{matrix} \right)\!\!\!\!\!\!}_{\text{$p$ times $\mathbf{A}$}} 
\end{equation}

\subsection{Extracting predictable single components} \label{sec:extractIsolated}
In section \ref{sec:pfa} we initially stated a prediction model that always scopes on single components.
This idea was not suitable for PFA because it contradicts the orthogonal agnosticity criterion.
In this section we propose a strategy to extract well predictable single components even though. We begin
by recalling our initial notion of linear auto regressive predictability:
\begin{align}
 \mathbf{a}^T \mathbf{z}(t) \quad \appr^! \;& \quad b_{1} \mathbf{a}^T \mathbf{z}(t-1) + \ldots + b_{p} \mathbf{a}^T \mathbf{z}(t-p) \label{pfa-criterionARScalar2}\\
= \;& \quad \mathbf{a}^T \hist_{\mathbf{z}, p}(t) \; \mathbf{b}
\end{align}
It is possible to write this for multiple dimensions by constraining the coefficient-matrices to be diagonal:
\begin{equation} \label{pfa-criterionARDiagMatrix2}
 \mathbf{m}(t) \quad \appr^! \quad \mathbf{B}_{1} \mathbf{m}(t-1) + \ldots + \mathbf{B}_{p} \mathbf{m}(t-p) \quad \text{with}\quad \mathbf{B}_i \in \mathbb{R}^{n \times n} \text{, diagonal}
\end{equation}
This model is not orthogonal agnostic, so a different approach than in section \ref{sec:pfa} is needed.
To minimize the least-squares-error of \eqref{pfa-criterionARScalar2}, the following optimization
problem needs to be solved:

\begin{align} \label{pfaOrigin}
	\begin{split} 
		\optmin{\mathbf{a} \in \mathbb{R}^n,\;\mathbf{b} \in \mathbb{R}^p} & \av{\left( \mathbf{a}^T (\mathbf{z} - \hist_{\mathbf{z}, p} \mathbf{b}) \right)^2 \;} \\
		\subjectto	& \mathbf{a}^T \av{\mathbf{z}} \hphantom{\mathbf{a} \mathbf{z}^T} \, \quad = \quad 0 \quad \quad \text{(zero mean)}\\
				& \mathbf{a}^T \av{\mathbf{z} \mathbf{z}^T} \mathbf{a} \quad = \quad 1 \quad \quad \text{(unit variance)}
	\end{split}
\end{align}
Via analytic optimization it is straight forward to find the optimal $\mathbf{a}$, if $\mathbf{b}$ is fixed and vice versa:

If $\mathbf{b}$ is fixed, choose $\mathbf{a}$ as the eigenvector corresponding to the smallest eigenvalue in
\begin{equation} \label{pfa_b_to_a}
	\bigav{\mathbf{z}\mathbf{z}^T} - \bigav{\mathbf{z}\mathbf{b}^T\hist_{\mathbf{z}, p}} - \bigav{\hist_{\mathbf{z}, p}^T\mathbf{b}\mathbf{z}^T} + \bigav{\hist_{\mathbf{z}, p}\mathbf{b}\mathbf{b}^T\hist_{\mathbf{z}, p}^T}
\end{equation}

If $\mathbf{a}$ is fixed, choose $\mathbf{b}$ as
\begin{equation} \label{pfa_a_to_b}
	\mathbf{b}^T \; \coloneq \; \bigav{\mathbf{z}^T\mathbf{a}\mathbf{a}^T\hist_{\mathbf{z}, p}} \bigav{\hist_{\mathbf{z}, p}^T\mathbf{a}\mathbf{a}^T\hist_{\mathbf{z}, p}}^{-1}
\end{equation}
By inserting \eqref{pfa_a_to_b} into \eqref{pfaOrigin} one could obtain a problem written in $\mathbf{a}$ only:
\begin{align} \label{pfaOriginConcerning_a}
	\begin{split} 
		\optmin{\mathbf{a} \in \mathbb{R}^n} & \av{\left( \mathbf{a}^T (\mathbf{z} - \hist_{\mathbf{z}, p} \bigav{\mathbf{z}^T\mathbf{a}\mathbf{a}^T\hist_{\mathbf{z}, p}} \bigav{\hist_{\mathbf{z}, p}^T\mathbf{a}\mathbf{a}^T\hist_{\mathbf{z}, p}}^{-1}) \right)^2 \;} \\
		\subjectto	& \mathbf{a}^T \av{\mathbf{z}} \hphantom{\mathbf{a} \mathbf{z}^T} \, \quad = \quad 0 \quad \quad \text{(zero mean)}\\
			& \mathbf{a}^T \av{\mathbf{z} \mathbf{z}^T} \mathbf{a} \quad = \quad 1 \quad \quad \text{(unit variance)}
	\end{split}
\end{align}
Problem \eqref{pfaOriginConcerning_a} is not efficiently globally solvable by any method known to us, which is mainly due to the occurrence of $\mathbf{a}$ in a matrix-term
under an inversion-symbol. However a possible strategy is to approximate the solution by choosing an initial value for $\mathbf{a}$ or $\mathbf{b}$ and
applying \eqref{pfa_b_to_a} and \eqref{pfa_a_to_b} in turns until a stable state is reached.

As a reasonable initial value for this procedure we choose $\mathbf{b}$ such that it is the best predictor of $\mathbf{z}$ on average, in absence of any $\mathbf{a}$:
\begin{equation} \label{averageBest_b}
	\mathbf{z}(t) \quad \appr^! \quad b_{1} \mathbf{z}(t-1) + \ldots + b_{p} \mathbf{z}(t-p) \quad = \quad \hist_{\mathbf{z}, p}(t) \mathbf{b}
\end{equation}
To minimize the error of \eqref{averageBest_b} on average over all components of $\mathbf{z}$, we propose the following least-squares optimization:

\begin{equation} \label{startB}
	\optmin{\mathbf{b} \in \mathbb{R}^p} \av{ (\mathbf{z} - \hist_{\mathbf{z}, p} \mathbf{b})^T (\mathbf{z} - \hist_{\mathbf{z}, p} \mathbf{b})}
\end{equation}
The solution of this problem is
\begin{equation} \label{startBSolution}
	\mathbf{b} \; \coloneq \; \bigav{\mathbf{z}^T\hist_{\mathbf{z}, p}} \bigav{\hist_{\mathbf{z}, p}^T\hist_{\mathbf{z}, p}}^{-1}
\end{equation}
Solution \eqref{startBSolution} does not change, if we replace $\mathbf{z}$ by $\mathbf{A}^T\mathbf{z}$ with any orthogonal, full ranked $\mathbf{A}$.
However, one quickly finds examples, where the procedure stabilizes in sub-optimal states. Though one can partly overcome this issue by
estimating better starting points, the method still has unknown success-probability.

Probably a better possibility is to solve \eqref{pfaOrigin} with PFA as described in section \ref{sec:extraction} for $r = 1$.

After extracting one component either way, one can project $\mathbf{z}$ to the signal space uncorrelated (i.e.\ orthogonal) to the extracted component. The extraction- and projection-procedure can be repeated until any desired number of components is extracted.

%

\subsection{Solving inhomogeneous eigenvalue problems} \label{sec:inhomEVP}
An inhomogeneous eigenvalue problem like \eqref{ipfaOptControlInhomEVP} in general form can be stated as
\begin{align} \label{inhomEVP}
\mathbf{A} \mathbf{v} \quad = \quad &\lambda \mathbf{v} + \mathbf{b} \\
\norm{\mathbf{v}} \quad = \quad &c \label{inhomEVPconstr}
\end{align}
with matrix $\mathbf{A}$, vectors $\mathbf{v}$ and $\mathbf{b}$, scalar $c$ of appropriate dimensions.
While \cite{MATTHEIJ1987507} mainly focuses on direct numerical approaches, they also point out a method to write \eqref{inhomEVP} as an ordinary eigenvalue problem of higher dimension.
Since inhomogeneous eigenvalues are different from homogeneous ones for $\mathbf{b} \neq 0$, we can assume that $(\mathbf{A} - \lambda \mathbf{I})$ is invertible. Thus
we can calculate $\mathbf{v}$ as
\begin{equation} \label{calcV}
	\mathbf{v} \quad = \quad (\mathbf{A} - \lambda \mathbf{I})^{-1} \mathbf{b}
\end{equation}
and only have to find the inhomogeneous eigenvalues $\lambda$. To do so, we insert \eqref{calcV} into \eqref{inhomEVPconstr} and obtain
\begin{equation} \label{schurComplementIEVP}
	\mathbf{b}^T (\mathbf{A} - \lambda \mathbf{I})^{-T} (\mathbf{A} - \lambda \mathbf{I})^{-1} \mathbf{b} \quad = \quad c
\end{equation}
which implies that the block matrix
\begin{equation} \label{blockMatrixIEVP}
\left( \begin{matrix} (\mathbf{A} - \lambda \mathbf{I})(\mathbf{A} - \lambda \mathbf{I})^T & \mathbf{b} \\ \mathbf{b}^T & c \end{matrix} \right)
\end{equation}
can't have full rank (\eqref{schurComplementIEVP} is the Schur complement of \eqref{blockMatrixIEVP}). Consequently its determinant must be zero. With the block matrix determinant formula
\begin{equation} \label{blockMatrixDet}
\det \left( \begin{matrix} \mathbf{A} & \mathbf{B} \\ \mathbf{C} & \mathbf{D} \end{matrix} \right) \quad = \quad \det(\mathbf{D}) \det(\mathbf{A} - \mathbf{B} \mathbf{D}^{-1} \mathbf{C})
\end{equation}
we can state this as
\begin{equation} \label{detBlockMatrixIEVP}
\det \Big( (\mathbf{A} - \lambda \mathbf{I})(\mathbf{A} - \lambda \mathbf{I})^T - \frac{\mathbf{b} \mathbf{b}^T}{c} \Big) \quad = \quad 0
\end{equation}
So $( (\mathbf{A} - \lambda \mathbf{I})(\mathbf{A} - \lambda \mathbf{I})^T - \frac{\mathbf{b} \mathbf{b}^T}{c} )$ must have at least one zero-valued eigenvalue and there exists a corresponding eigenvector:
\begin{equation} \label{detBlockMatrixNonzeroEV}
\exists \; \mathbf{x} \neq 0 \colon \qquad \lambda^2 \mathbf{x} + \lambda (-\mathbf{A}^T - \mathbf{A}) \mathbf{x} + \Big(\mathbf{A}\mathbf{A}^T - \frac{\mathbf{b} \mathbf{b}^T}{c} \Big) \mathbf{x} \quad = \quad 0
\end{equation}
\eqref{detBlockMatrixNonzeroEV} can be written as
\begin{equation} \label{blockMatrixIEVPFinal}
\exists \; \mathbf{x} \neq 0 \colon \qquad \left( \begin{matrix} 0 & \mathbf{I} \\ \frac{\mathbf{b} \mathbf{b}^T}{c} - \mathbf{A}\mathbf{A}^T  & \mathbf{A}^T + \mathbf{A} \end{matrix} \right) \; \left( \begin{matrix} \mathbf{x} \\ \lambda \mathbf{x} \end{matrix} \right) \quad = \quad \lambda \; \left( \begin{matrix} \mathbf{x} \\ \lambda \mathbf{x} \end{matrix} \right)
\end{equation}
Finally we can use any ordinary eigenvalue algorithm to obtain $\lambda$ as an eigenvalue of
$\left( \begin{smallmatrix} 0 & \mathbf{I} \\ \frac{\mathbf{b} \mathbf{b}^T}{c} - \mathbf{A}\mathbf{A}^T  & \mathbf{A}^T + \mathbf{A} \end{smallmatrix} \right)$.
\end{document}